\documentclass{iopjournal}

\usepackage{graphicx}
\usepackage{amsmath}
\usepackage{amssymb}
\usepackage{amsfonts}
\usepackage{booktabs}
\usepackage{bm}
\usepackage{listings}
\usepackage{xcolor}
\usepackage{algpseudocode}
\usepackage{algorithm}
\usepackage{tabularx}
\usepackage{wrapfig}
\usepackage{caption}
\usepackage{subcaption}
\usepackage{makecell}
\usepackage[normalem]{ulem} 
\usepackage{cancel} 
\usepackage{algpseudocode}
\usepackage{siunitx}
\usepackage{orcidlink}

\definecolor{federicoColor}{rgb}{0.0,0.0,0.75}

\usepackage[%
backend=biber,%
bibstyle=numeric,
citestyle=numeric-verb,
hyperref=true,%
backref=false,%
sorting=none,%
maxnames=99,%
isbn=false,%
block=ragged,
url=false,%
]{biblatex}
\makeatletter
\newcommand{\LineComment}[1]{\Statex \hskip\ALG@thistlm \textit{// #1}}
\makeatother

\begin{document}
\articletype{paper} \\
\title{Adaptive-Frequency Resonate-and-Fire Neurons for Spectral Estimation of Streaming Radar Signals}
\author{ Stefano Chiavazza$^{1,*}$\orcidlink{0009-0003-0854-685X}, Sen Yuan$^2$\orcidlink{0000-0003-0175-6662}, Marc Geilen$^{1}$\orcidlink{0000-0002-2629-3249}, Francesco Fioranelli$^2$\orcidlink{0000-0001-8254-8093}, Federico Corradi$^{1}$\orcidlink{0000-0002-5868-8077}
	\\
    $^1$ Electronic Systems, Eindhoven University of Technology, The Netherlands \\
	  $^2$ MS3 Group, Delft University of Technology, The Netherlands \\
	$^*$ Author to whom any correspondence should be addressed.}


\begin{abstract}
    Frequency Modulated Continuous Wave (FMCW) radar systems traditionally rely on Fourier-based methods, such as the Fast Fourier Transform (FFT), to estimate target range and velocity. While computationally efficient, these approaches require storing and processing large blocks of data, which can become a bottleneck in memory-constrained or low-latency applications. In this work, we propose a neuromorphic-inspired signal processing method based on adaptive resonate-and-fire (ARF) neurons formulated as a discrete-time dynamical system. Each neuron dynamically adjusts its internal frequency to match dominant frequency components of the input radar signal, enabling direct estimation of target ranges and velocities without computing the full frequency spectrum. The proposed model operates in a sample-by-sample fashion, resulting in memory requirements that scale with the number of tracked targets rather than the signal length. A feedback mechanism is also introduced to enable multiple neurons to lock on distinct frequency components in multi-target cases.
    Results on simulated and experimental data demonstrate that the method can successfully track multiple targets. Compared to conventional FFT-based approaches, the proposed method offers reduced memory usage proportional only to the number of tracked targets, making it suitable for resource-constrained and edge-based radar applications.
\end{abstract}

\section{Introduction}

    \begin{figure}[t]
        \includegraphics[width=\textwidth]{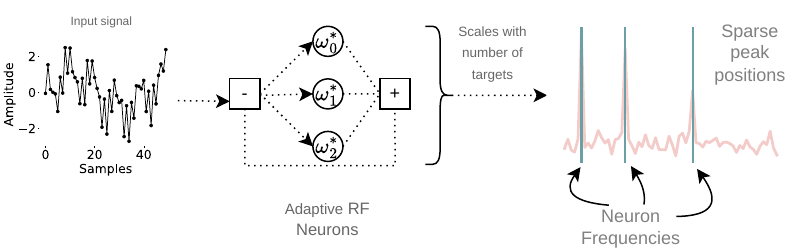}
        \caption{Working principle of the adaptive resonate-and-fire (ARF) network proposed for this work. Each ARF neuron receives a one-dimensional signal sample-by-sample and, through online frequency adaptation and mean-field feedback, converges onto the tuning frequency of a distinct spectral peak. This same primitive is applied twice in the FMCW radar processing pipeline: on the fast-time signal for range estimation, and on the slow-time signal for Doppler/velocity estimation. More generally, this can substitute any sparse-FFT stage in which only a few dominant frequencies matter. Notably, the neuron count scales with the number of targets rather than with the resolution in range and Doppler.}
        \label{fig:scheme_working_principle}
    \end{figure}

    Frequency-Modulated Continuous-Wave (FMCW) radar is a key sensing modality for automotive, robotics, and short-range presence-detection applications, thanks to its capabilities to remain reliable under adverse weather and lighting conditions that degrade cameras and LiDAR~\cite{zhang_perception_2023,engels_automotive_2021}. Modern mmWave radars deliver increasingly fine range--Doppler resolution, but the resulting data volumes must be processed in real time on embedded platforms with tight energy budgets, leading to bottlenecks~\cite{engels_automotive_2021,horowitz_computings_2014}. 
    In these cases, the conventional processing pipeline performs a two-dimensional Fast Fourier Transform (FFT) along fast and slow time, followed by Constant False Alarm Rate (CFAR) detection~\cite{rohling_ordered_2011}: a frame-based procedure whose memory footprint and off-chip bandwidth scale with the full range--Doppler grid,  and thus become the dominant bottleneck at the desired high resolutions~\cite{akin_memory_2012,kortli_hwsw_2022,harishore_singh_energy-efficient_2024,leitersdorf_fourierpim_2023}.

    Neuromorphic computing offers a complementary, event-driven path to low-power radar processing by exploiting sparse activity and in-memory state~\cite{davies_advancing_2021,indiveri_memory_2015}. Two families of approaches dominate the radar literature. The first replaces the Discrete Fourier Transform operations with spiking circuits, either through rate coding~\cite{lopez-randulfe_spiking_2021}, time-to-first-spike (TTFS) encoding~\cite{lopez-randulfe_time-coded_2022,wang_snn-ft_2025}, or chirp-by-chirp 2D-FFT spike pipelines~\cite{chiavazza_low-latency_2025}. The second uses resonate-and-fire (RF) neurons~\cite{noauthor_resonate-and-fire_nodate,orchard_efficient_2021} whose fixed tuning frequencies sample the spectrum directly, as in RF-based interference detection~\cite{hille_resonate-and-fire_2022,shaaban_resonate-and-fire_2024}, the NeuroRadar IoT sensor~\cite{zheng_neuroradar_2023}, and the recently proposed Spiking Neural Resonator (SpiNR) for range--angle~\cite{reeb_range_2025} and range--velocity estimation with Activity-Gated Sparsity (AGS) on Intel Loihi~2~\cite{reeb_energy-efficient_2026}. Across both families, however, the number of neurons is tied to the spectrum resolution: $\mathcal{O}(L)$ or $\mathcal{O}(L\log L)$ for spiking FFTs, where $L$ is the signal length.
    Efficiency gains come from coding schemes or post-hoc activity gating, not from changing the underlying representation. For a complete review on neuromorphic radar signal analysis and classification we refer the reader to the work of  Hamrell et al. in~\cite{Hamrell_2026}.

    In this work, we propose a novel route that is \emph{adaptive}, \emph{peak-finding}, and \emph{per-target} (Fig.~\ref{fig:scheme_working_principle}). Specifically, we develop a discrete-time Adaptive Resonate-and-Fire (ARF) neuron, built on the adaptive-frequency oscillators of Righetti and Buchli~\cite{righetti_dynamic_2006,buchli_frequency_2008}, whose internal frequency is learned online from the radar signal rather than pre-configured. 
    Because each neuron adapts its frequency, it converges onto a spectral \emph{peak} rather than estimating a single fixed frequency, so the network does not need to compute the full spectrum, and a separate CFAR/detection stage becomes unnecessary. A mean-field feedback mechanism forces distinct neurons onto distinct targets. The resulting representation is structurally sparse: memory scales with the number of physical reflectors $K$, not with the range-Doppler grid size $L_r\cdot L_v$ ($L_r$ refers to number of range bins and $L_v$ refers to the number of velocity bins), and is therefore typically orders of magnitude smaller in realistic automotive or indoor scenes.

    The main contributions of this paper are: (i)~an online-adaptive discrete-time ARF neuron model that learns its tuning frequency from the raw baseband signal; (ii)~a fully streaming, sample-by-sample range-Doppler processing pipeline requiring no chirp or frame buffer on either processing layer; (iii)~direct target detection without a dedicated CFAR stage, since each converged neuron already localizes a peak; (iv)~memory complexity $\mathcal{O}(K)$ set by the number of tracked targets rather than by the spectrum resolution; 
    and (v)~a bandwidth-efficient spike output that emits events only when a neuron's estimated frequency changes by more than a user-defined threshold.

    We validate the proposed approach on both synthetic FMCW data and recorded data covering single-target and multi-target scenes, and compare computational and memory complexity against conventional FFT baselines and state-of-the-art neuromorphic radar pipelines. This work focuses on the algorithmic contribution: the sample-by-sample, state-based formulation of the ARF model is deliberately designed to map onto neuromorphic hardware, which we will target in the future. The remainder of the paper is organized as follows: Section~\ref{sec:related} positions the proposed approach against the existing literature; Section~\ref{sec:methods} introduces the ARF model and the feedback and spike-generation mechanisms; Section~\ref{sec:results} reports simulation and recorded-data experiments; the final section discusses complexity and directions for hardware implementation.

\section{Related Works}
\label{sec:related}

    Three distinct research lines shape the context of this work, and we review them in turn. Section~\ref{sec:rw_fft} covers the conventional FFT + CFAR pipeline and the memory bottlenecks that motivate alternatives; Section~\ref{sec:rw_neuro} covers neuromorphic radar processing, which attacks the same bottlenecks but still allocates neurons per individual spectral bin; and Section~\ref{sec:rw_osc} covers the dynamical-systems literature on adaptive-frequency oscillators, from which we draw the key mechanism of our model.

    \subsection{FFT-based radar pipelines and memory bottlenecks}
    \label{sec:rw_fft}
    In general, all conventional FMCW radar systems in production stages rely on a two-dimensional FFT followed by a CFAR detector~\cite{engels_automotive_2021,rohling_ordered_2011}. Although the FFT itself is $\mathcal{O}(L\log L)$, where $L$ is the number of sample, with respect to its complexity, practical implementations on FPGAs, ASICs, and general-purpose accelerators suffer by off-chip memory accesses to the range--Doppler data cube~\cite{akin_memory_2012,kortli_hwsw_2022}. A recurring theme in the recent literature is the pursuit of efficiency through sparsity: sparse FFT algorithms~\cite{hassanieh_simple_2012} and energy-efficient sparse FFT accelerators with compress transpose memory~\cite{harishore_singh_energy-efficient_2024} exploit the fact that only a small fraction of spectral bins contain targets, while processing-in-memory FFTs seeks to close the memory-bandwidth gap at the device level~\cite{leitersdorf_fourierpim_2023}. 
    All of these approaches nevertheless commit to first computing (or allocating storage for) the full spectrum, and only afterwards discarding the empty bins.

    \subsection{Neuromorphic radar: DFT/FFT replacements, fixed resonators, and AGS}
    \label{sec:rw_neuro}
    Neuromorphic implementations of radar signal processing follow the same efficiency intuition but move it into spiking hardware. The rate-coded spiking DFT of L\'opez-Randulfe \textit{et al.}~\cite{lopez-randulfe_spiking_2021} and its subsequent TTFS variants~\cite{lopez-randulfe_time-coded_2022,wang_snn-ft_2025} reproduce the Fourier transform as a feed-forward spiking network with $\mathcal{O}(L)$ or $\mathcal{O}(L\log L)$ neurons. RF-based approaches instead exploit the temporal dynamics of resonator neurons~\cite{noauthor_resonate-and-fire_nodate,orchard_efficient_2021} to analyze the radar signal directly in the time domain, with applications to interference detection~\cite{hille_resonate-and-fire_2022}, gesture and target recognition~\cite{shaaban_resonate-and-fire_2024}, low-power IoT sensing~\cite{zheng_neuroradar_2023}, and chirp-by-chirp 2D-FFT range--Doppler estimation~\cite{chiavazza_low-latency_2025}. The Spiking Neural Resonator (SpiNR)~\cite{reeb_range_2025} refines this approach by assigning one resonator per range (and, in its most recent extension~\cite{reeb_energy-efficient_2026}, per range--velocity) bin, coupling it with a spiking Ordered-Statistics CFAR detector and an Activity-Gated Sparsity (AGS) mechanism that switches off non-resonating neurons on Intel Loihi~2. SpiNR--AGS is, to our knowledge, the most efficient reported neuromorphic radar pipeline to date; critically, however, its efficiency is obtained by \emph{gating} bins rather than removing them: the $\mathcal{O}(L_r\cdot L_v)$ neuronal population is still instantiated, and AGS then suppresses activity in the bins that carry no target.

    \subsection{Adaptive oscillators}
    \label{sec:rw_osc}
    An orthogonal line of work, originating outside the radar community, studies dynamical systems whose oscillators \emph{learn} their own natural frequency from a driving input. Righetti, Buchli, and Ijspeert introduced dynamic Hebbian learning for adaptive-frequency Hopf oscillators~\cite{righetti_dynamic_2006} and showed that a negatively mean-field coupled pool of such oscillators performs online frequency analysis~\cite{buchli_frequency_2008}, a result with deep connections to phase-frequency synchronization of globally coupled populations~\cite{acebron_adaptive_1998}. A similar model has also been used to explain the observed data on synchrony in fireflies \cite{ermentrout_adaptive_1991}. The adaptive-oscillator paradigm has since been extended to Duffing dynamics~\cite{perkins_duffing_2025,yang_vibrational_2026} and to physical reservoir computing~\cite{shougat_self-learning_2024}, and a dedicated VLSI implementation of the adaptive Hopf oscillator has been demonstrated~\cite{ahmadi_vlsi_2011}. 
    
    In parallel, analog neuromorphic substrates are increasingly recognized as a natural substrate for RF dynamics: early analog VLSI resonate-and-fire neurons~\cite{nakada_analog_2006}, direct signal encoding with analog RF neurons in modern CMOS~\cite{lehmann_direct_2023}, and the very recent 22 \, nm linear analog implementation of a resonate-and-fire neuron of Liu \textit{et al.}~\cite{liu_linear_2025} suggest that continuous-time oscillatory dynamics can be realized on analog hardware.

	\vspace{0.2cm}	
    In summary, in contrast to all approaches above, we propose to replace a bin-indexed set of fixed-frequency resonators with a \emph{sparse} set of adaptive oscillatory neurons, whose count scales with the number of physical reflectors rather than with the range--Doppler grid size. The neurons produce peak estimates directly in a single streaming pass and remove the need for a separate spectral-allocation or CFAR-based detection stage, as depicted in Fig.~\ref{fig:scheme_working_principle}.

\section{Methods}
\label{sec:methods}

In this section, we present the adaptive resonance-and-fire (ARF) model for radar signal processing.

Buchli \textit{et al.}~\cite{buchli_frequency_2008} use a continuous-time dynamical system to implement an oscillator with adaptive frequency. Here, we present an analogous discrete-time formulation. A single neuron/oscillator with state variables $z_t$ and $\omega_t$ is updated according to:

\begin{align}
    \omega_{t} &= \omega_{t-1} - \alpha\lambda \left( \frac{\operatorname{Im}(z_t)}{|z_t|} \operatorname{Re}(I_t) - \frac{\operatorname{Re}(z_t)}{|z_t|} \operatorname{Im}(I_t) \right) \nonumber \\
    z_{t+1} &= (z_t + I_t) \cdot \exp{i \omega_{t} \Delta t}, 
    \label{eq:base_adaptive}
\end{align}

where $z_t$ is the neuron state with complex values and $\omega_t$ its instantaneous frequency with real values at time $t$, $I_t$ is the input signal with complex values, $\alpha$ is a scaling factor and $\lambda$ is the learning rate for frequency update. For a single neuron/oscillator, $I_t$ coincides with the raw radar signal $F_t$, i.e., the complex-valued IQ signal digitized by an FMCW radar receiver.
Notably, when several neurons are coupled through the feedback mechanism introduced later in Eq.~\eqref{eq:mean_field}, $I_t$ denotes the feedback-corrected driving signal, while $F_t$ remains the uncorrected radar signal used in the frequency-update correlation.

The first term in Eq. \ref{eq:base_adaptive} represents simply a harmonic oscillator with frequency $\omega$. The second term measures the phase difference between the neuron and the input signal and adjust the neuron's frequency accordingly. Intuitively, for an input signal with a single frequency component, the neuron will adjust its frequency to match that of the input. The frequency update is determined by the phase difference between the neuron and the input signal. Over time, the frequency of the neuron will evolve and match the frequency of the input. The update rule resembles a phase-locked loop (PLL), but unlike a PLL the neuron state is also driven by the input, since $I_t$ is added to the state at every timestep.

The role of the scaling factor $\alpha$ is to scale the measured phase difference to the desired frequency range. Therefore, the choice of $\alpha$ is related to the sampling frequency of the signal. We set $\alpha = 100 f_s$ in this work, where $f_s$ denotes the sampling frequency. The scaling is further refined by adjusting the learning rate $\lambda$.

\subsection{Feedback Mechanism for Neurons at Different Frequencies}

If multiple frequency components are present in the input signal (as in the case of multiple targets for an FMCW radar), multiple neurons are needed to cover all frequencies. However, without other constraints, all the neurons are likely to converge to the same frequency. In order to prevent this, the work in \cite{buchli_frequency_2008} proposes a mean-field feedback mechanism that pushes the oscillators to converge to different frequencies. Here, we propose an analogous discrete-time version of the feedback mechanism.

\begin{figure}[b]
    \centering
    \includegraphics[width=.6\linewidth]{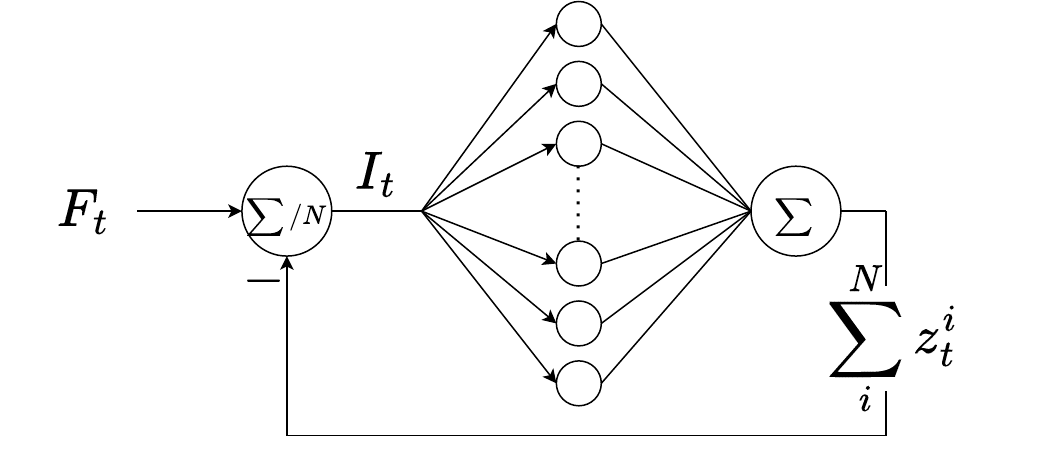}
    \caption{Mean-field feedback connections scheme. The output signals of all the neurons are summed together and subtracted from the input. The goal is to cancel out from the input signal frequencies that are already covered by a neuron.}
    \label{fig:feedback_scheme}
\end{figure}

Let $N$ be the number of adaptive-frequency neurons as described in equation \eqref{eq:base_adaptive}. Then the input signal becomes:

\begin{align}
    I_t = \frac{1}{N} \left( F_t -  \sum_{i=1}^{N} z_{t}^i \right)
    \label{eq:mean_field}
\end{align}

where $F_t$ is the reference signal, i.e., the uncorrected radar signal, and the feedback signal is the sum of the signals produced by each neuron. The scaling of $\frac{1}{N}$ helps to ensure stability. This effectively creates a negative feedback loop that is clearly shown in the connection scheme in figure \ref{fig:feedback_scheme}.

The specific implementation of the feedback mechanism is explained in Algorithm~\ref{alg:neuron_update}. Intuitively, the signal generated by one neuron is subtracted from the input signal. As a result, the other neurons will not be affected by the frequency components already covered by other neurons. The neurons start with a random frequency to break symmetry and, as more samples are processed, they will adapt their frequencies to match the input. Once all the neurons have converged, the neurons are effectively reconstructing the input signal. Assuming that no noise is present, the resulting signal after the negative feedback has been applied will be zero.

\begin{figure}
    \begin{subfigure}{0.49\textwidth}
        \centering
        \includegraphics[width=\linewidth]{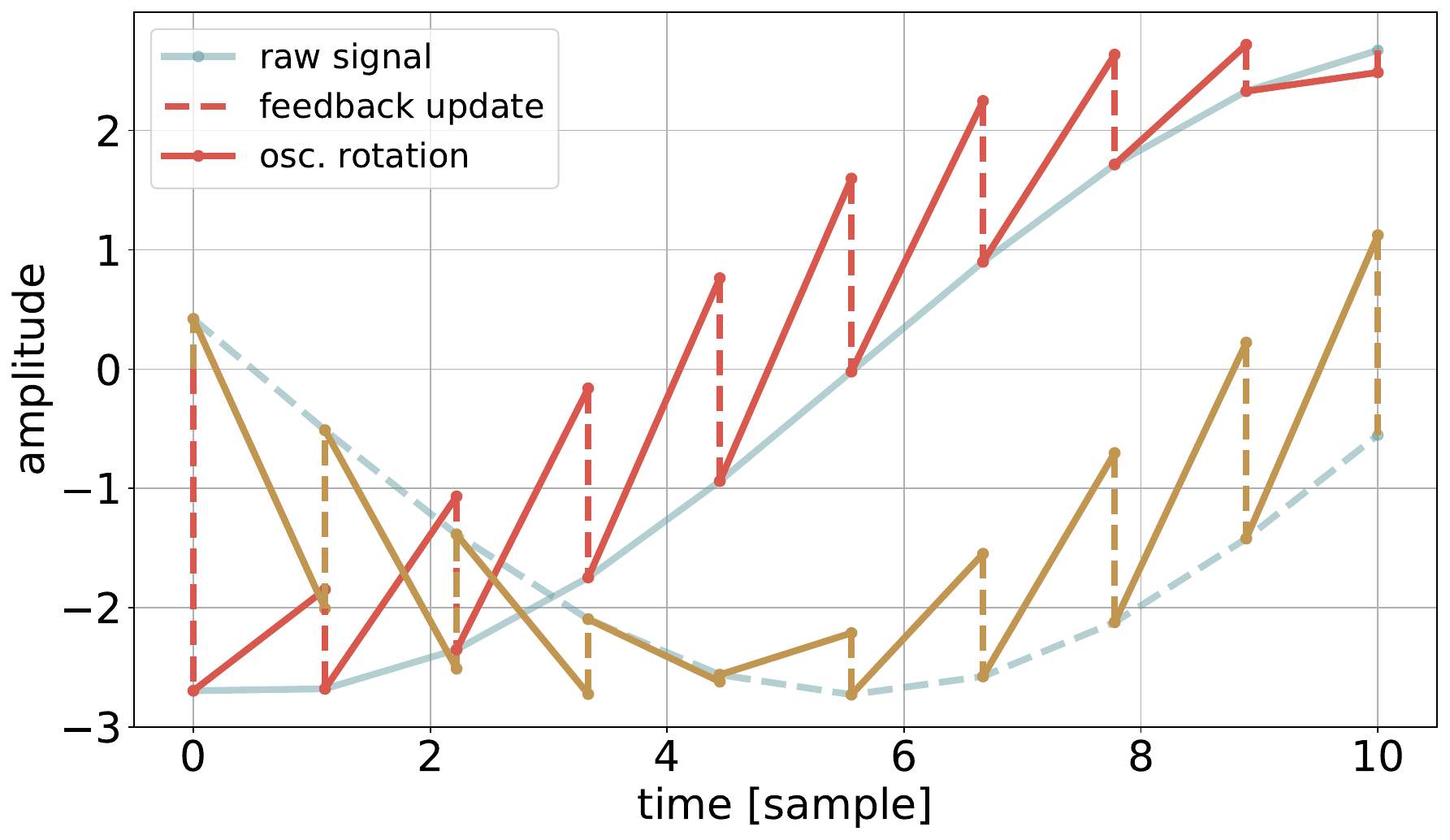}
        \caption{Beginning of the simulation.}
        \label{fig:system_dynamics_beginning}
    \end{subfigure}
    \begin{subfigure}{0.49\textwidth}
        \centering
        \includegraphics[width=\linewidth]{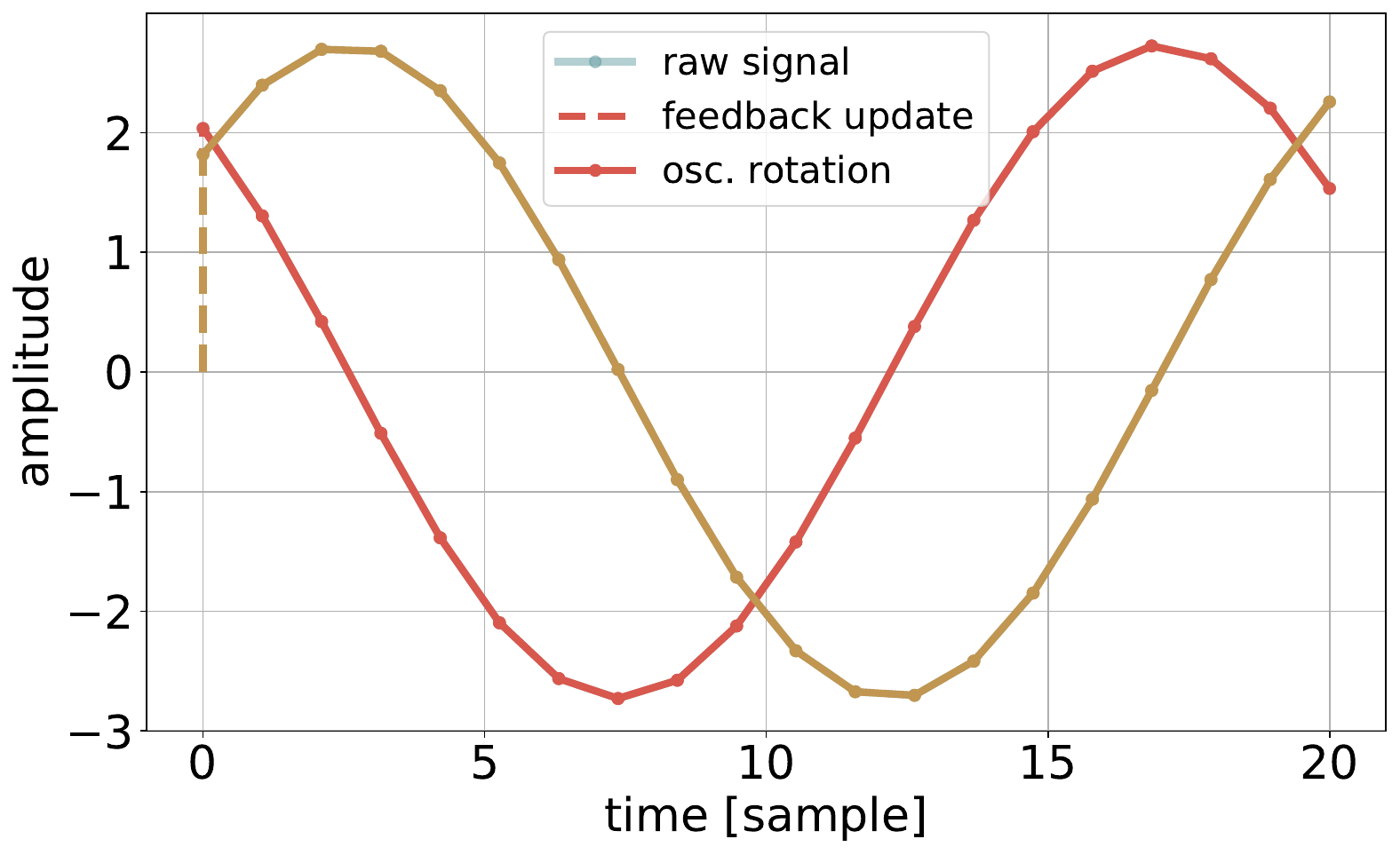}
        \caption{End of the simulation.}
        \label{fig:system_dynamics_end}
    \end{subfigure}

    \caption{System dynamics during the simulation. The red and yellow lines represent the real and imaginary components respectively. At the beginning of the simulation, the neuron frequency is random and does not match the input. The solid lines show the oscillation caused by the neuron frequency, and the dotted lines show the effect of the feedback mechanism. Over time the neuron frequency will converge to the signal's frequency and the neuron rotation will match the input signal exactly.}
    \label{fig:system_dynamics}
\end{figure}

Due to the nonlinear dynamical nature of the system, a closed-form solution describing its behavior could not be derived. However, we can analyze the simpler case with a single neuron.

Figure~\ref{fig:system_dynamics_beginning} illustrates the effect of feedback in the case of a single tone input frequency and a single neuron. The neuron starts with a random frequency different from the signal's frequency and uses it to apply a rotation to the neuron's state; these are the solid lines in the figure. 
After the frequency update based on the phase error, the feedback signal is added to the neuron's state. In the case of a single neuron, this action simplifies to: $$z_{t+1}' = z_t + I = z_t + F - z_t = F$$ 
meaning that the neuron state is reset exactly to the most recent input value. This only applies to the single-neuron case, as in the general case this includes the scaling factor $1/N$. This guarantees that the phase difference measured in each step depends only on the frequency difference accumulated in a single step. 

Assuming that both the signal and the neuron have the same initial state $s$, the next sample of the input signal will correspond to: $F = s \cdot \exp(i f \Delta t)$, and the next neuron state will correspond to $z = s \cdot \exp(i \omega \Delta t)$. We can rewrite the neuron frequency as $\omega=f+\Delta f$. Since the state of the neuron and the input signal are always set to the same value by the action of the feedback mechanism, we can view the neuron's state as simply a phase shifted copy of the input signal. This phase shift corresponds to the frequency difference: $\Delta \phi = \Delta f \Delta t$.

Thus, we can rewrite the frequency update term in Eq. \ref{eq:base_adaptive} using this interpretation:
\begin{align}
    \Delta \omega &= -\alpha\lambda\left( \frac{\operatorname{Im}(z)}{|z|}\operatorname{Re}(I)
                - \frac{\operatorname{Re}(z)}{|z|}\operatorname{Im}(I) \right) \notag \notag\\
    &= -\alpha\lambda\frac{1}{|z|}\left[\operatorname{Im}(z)\left(\operatorname{Re}(F)-\operatorname{Re}(z)\right)
                -\operatorname{Re}(z)\left(\operatorname{Im}(F)-\operatorname{Im}(z)\right) \right] \notag\\
    &= -\alpha\lambda\frac{1}{|z|}\left(\operatorname{Im}(z)\operatorname{Re}(F)
                -\operatorname{Re}(z)\operatorname{Im}(F) \right) \notag\\
    &= -\alpha\lambda\frac{1}{|z|}[\left(\operatorname{Re}(F)\sin(\Delta \phi) + \operatorname{Im}(F)\cos(\Delta \phi)\right) \cdot \operatorname{Re}(F) \notag \\
    &\quad - \left(\operatorname{Re}(F)\cos(\Delta \phi) - \operatorname{Im}(F)\sin(\Delta \phi)\right) \cdot \operatorname{Im}(F) ] \notag \\
    &= -\alpha\lambda \frac{1}{|z|}\left( \operatorname{Re}(F)^2  + \operatorname{Im}(F)^2\right) \sin(\Delta \phi) \notag \\
    &= -\alpha\lambda |z|\sin(\Delta \phi)
    \label{eq:wdot_complex}
\end{align}
where $|z|$ is the absolute amplitude of the input signal. Therefore, in the case of a single neuron, its frequency evolution can be predicted by Eq. \ref{eq:wdot_complex}. Over time, the frequency of the neuron will evolve to match the frequency of the input. As shown in Figure \ref{fig:system_dynamics_end}, once the frequency convergence is reached, the oscillation perfectly overlaps with the input signal, causing the frequency update to drop to zero.

Note that the method also works in the case of real-only input signals $F_t$. In this case, the second term of the frequency update rule will be zero as the input has no imaginary component. This has an impact on the frequency convergence, but the overall behavior of the system is maintained. For this work we will only focus on complex-valued inputs for simplicity and because IQ sampling is common in radar processing.


Importantly, in the context of Frequency-Modulated Continuous-Wave (FMCW) radar processing, each frequency component of the dechirped intermediate frequency (IF) signal directly corresponds to a target at a distinct range \cite{RadarMagazine}. These target distances can be extracted by estimating the signal's frequency components, given the known parameters of the radar waveform—specifically, the chirp bandwidth and duration. 

An FMCW radar frame typically consists of multiple sequential chirps. To process these data with our proposed network, we concatenate these chirps into a single, continuous one-dimensional signal. However, concatenation introduces phase discontinuities at the chirp boundaries, which distorts the resonant frequency response of the neurons. To mitigate this and smooth the transitions, we applied a Hanning window to each individual chirp prior to concatenation. Consequently, by leveraging adaptive resonate-and-fire (ARF) neurons alongside the proposed feedback mechanism, each ARF neuron can successfully lock onto and track a single target, allowing target distances to be decoded directly from the steady-state firing frequencies of the neurons.

\begin{algorithm}[t]
\caption{Steps for the ARF update with feedback}
\label{alg:neuron_update}
\begin{algorithmic}[1]
\For{each sample $F$}
    \State $feedback \gets 0$
    \For{each neuron $n$}
        \LineComment{the notation $n.z$ denotes the variable $z$ belonging to neuron $n$}
        \State $feedback \gets feedback + n.z$
    \EndFor
    \LineComment{Calculate corrected input}
    \State $I \gets (F - feedback) / N$

    \For{each neuron $n$}
        \LineComment{Calculate phase difference according to Eq. \ref{eq:base_adaptive}}
        \State $\phi_{\Delta} \gets \operatorname{phaseDifference(I, n.z)}$ \label{alg:omega_delta_line}

        \LineComment{the notation $n.\omega$ denotes the variable $\omega$ belonging to neuron $n$} 
        \LineComment{Update neuron frequency: $n.\omega$}
        \State $n.\omega \gets n.\omega - \alpha\lambda \cdot \phi_{\Delta}$

        \LineComment{Add Input}
        \State $n.z \gets n.z + I$

        \LineComment{Apply rotation}
        \State $n.z \gets n.z \cdot \exp(i \cdot n.\omega \cdot dt)$
    \EndFor
\EndFor
\end{algorithmic}
\end{algorithm}

\subsection{Doppler/Velocity Estimation}

FMCW radars also allow to estimate the velocity of the targets via frequency processing \cite{RadarMagazine}. As mentioned in the previous sub-section, the distance of the targets is proportional to the frequency of the raw radar signal and can be extracted via an FFT operation on each single chirp.
A moving target will cause a small frequency shift in the signal as its distance changes over time. However, this frequency shift is usually too small to be reliably measured, as the target velocity is typically much slower than the duration of the chirps.
However, the target movement will also cause a phase shift in the signal that can be more easily measured and used to estimate the target's velocity via Doppler processing. In other words, each chirp will be phase shifted compared to the previous one depending on the target velocity. Given the cyclical nature of the phase, and looking at multiple chirps sequentially after each other, this phase evolution will generate a periodic signal whose frequency depends on the velocity of the targets. In conventional approaches, a second FFT operation is performed across multiple chirps (often referred to as the ``slow-time'' dimension) to estimate this Doppler frequency shift, and from there the target velocity. 

\begin{figure}
    \includegraphics[width=\textwidth]{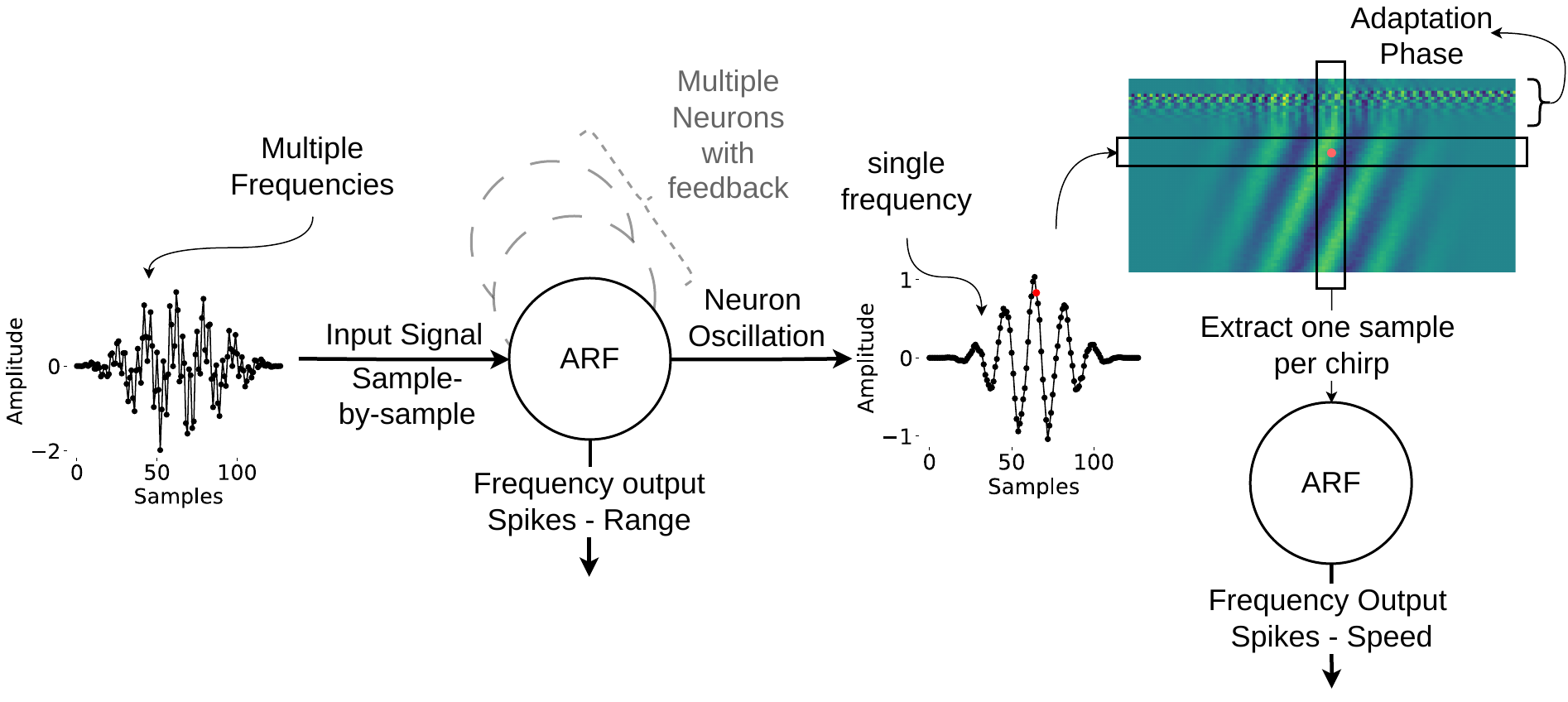}
    \caption{Processing pipeline for concurrent range and velocity estimation using two layers of adaptive resonate-and-fire neurons. The \textit{first-layer ARF neuron} isolates a single frequency component from the raw FMCW signal (corresponding to target range), and generates sparse spike outputs upon frequency convergence or tracking changes. Target velocity induces a slow-time phase shift across consecutive chirps, captured by extracting exactly one sample per chirp from the neuron's complex state $z$. The resulting phase-shifted sequence is fed directly into a \textit{second-stage ARF neuron} to resolve the target's Doppler velocity. Crucially, the entire pipeline operates sample-by-sample without requiring memory buffering. Information is maintained and updated entirely within the internal state dynamics of the neurons. Both the input signal and the neuron's oscillation are complex-valued; for clarity, we visualize only the real component.}
    \label{fig:doppler_scheme_single_neuron}
\end{figure}

\begin{figure}
    \includegraphics[width=\linewidth]{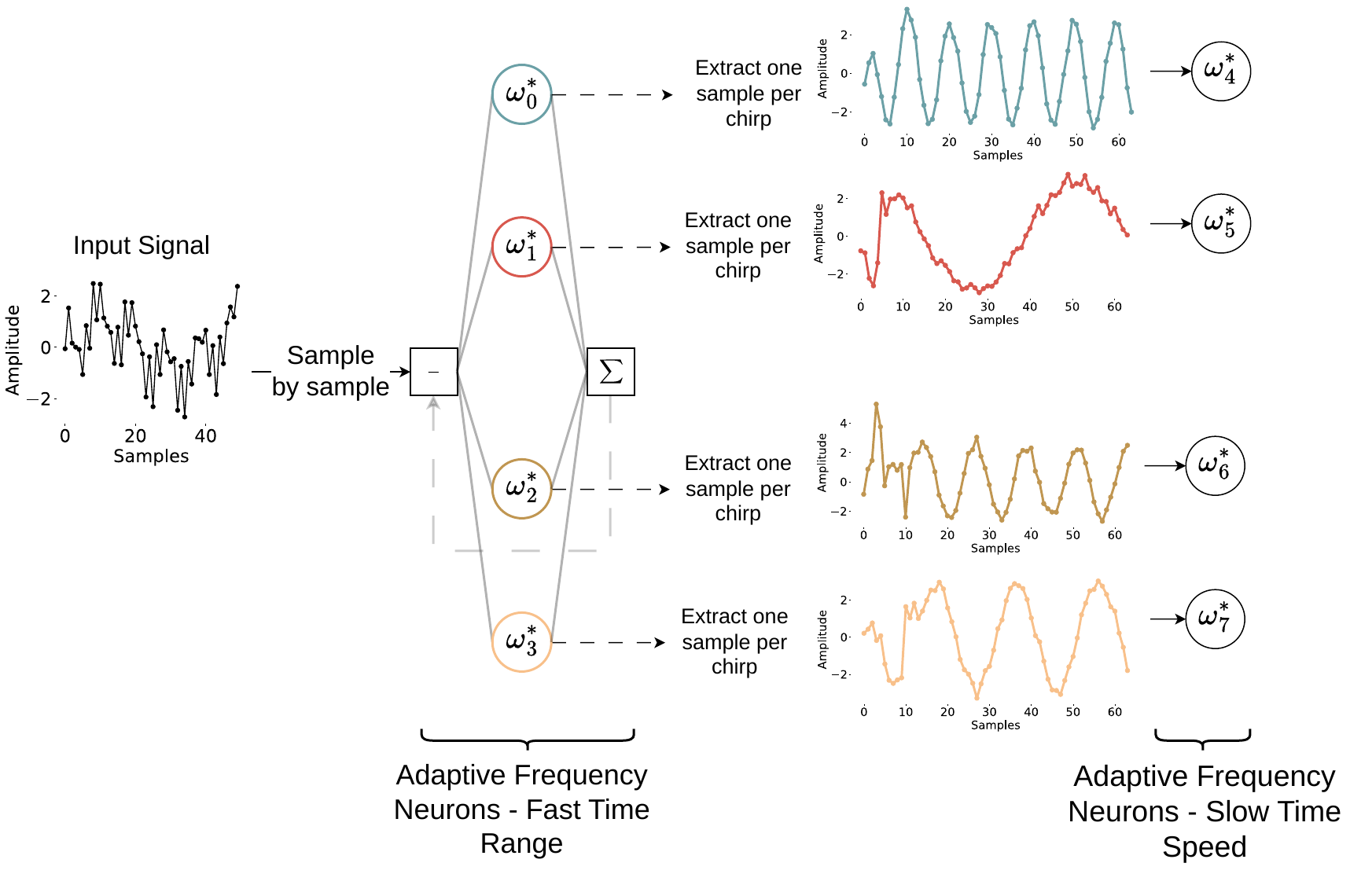}
    \caption{Complete processing pipeline for an example with 4 moving targets, showing the connection scheme for all the neurons in both range and velocity estimation layers.}
    \label{fig:doppler_scheme}
\end{figure}

In this work, we propose to use a second layer of adaptive resonate-and-fire (ARF) neurons to process this velocity signal. Figure \ref{fig:doppler_scheme_single_neuron} illustrates the connection scheme for extracting both range and velocity of a single target. Figure \ref{fig:doppler_scheme} shows the generalization to a multi-target case, with the full network architecture for the extraction of four targets. Within this network, the first layer of ARF neurons performs frequency extraction on the signal coming from the radar. As detailed in the previous subsection, this first stage corresponds to the range estimation processing.
After convergence, each neuron in the first layer isolates a single target. The complex state of each neuron then contains a single frequency component, with its phase evolving according to the target's movement. Figure \ref{fig:doppler_scheme_single_neuron} illustrates this phase change induced by the target's velocity. In this figure, the complex state of the neuron is reshaped such that each row corresponds to an individual chirp. By tracking the same sample index across all chirps (i.e., along a column of the reshaped matrix), we can analyze the phase evolution of the signal and consequently determine the target's velocity. 

To achieve this, a single adaptive neuron in the second layer processes this slow-time signal extracted from its corresponding first-layer neuron. Through this mechanism, the second-layer neurons converge directly to the Doppler frequencies related to the target velocities. Crucially, no data buffering is required at any stage of this pipeline. The first layer processes the radar signal sample-by-sample, and a single sample from each chirp is immediately forwarded to the next layer for processing. Consequently, there is no need to store the entire radar frame; Figure \ref{fig:doppler_scheme_single_neuron} includes the full frame solely for illustrative purposes.

\subsection{Spike Output Generation}

We have demonstrated how a bank of adaptive resonate-and-fire (ARF) neurons can be utilized for frequency analysis. However, continuously transmitting the instantaneous frequencies of all neurons at every timestep is inefficient with respect to bandwidth and data rate. To address this limitation, we propose an event-based encoding method that generates spikes only when the frequency shift of a neuron exceeds a predefined threshold.

This mechanism is implemented by maintaining a secondary, approximate frequency estimate, $\hat{\omega}$, for each neuron. Initially, the actual frequency $\omega$ and this approximate estimate are identical. As the system evolves, the actual frequency updates continuously, whereas the approximate value $\hat{\omega}$ is updated only when the absolute difference between the two exceeds a threshold $\theta$. When this condition is met, the neuron emits a spike, and $\hat{\omega}$ is immediately set to match the current actual frequency. This procedure is formalized in Algorithm~\ref{alg:spike_generation}. Each generated spike carries a polarity bit (positive or negative) indicating whether the tracked frequency is increasing or decreasing. For range and velocity estimation, the threshold $\theta$ can be defined based on a desired precision level. For instance, given a range tolerance of \SI{5}{\cm} and a set of known radar parameters, the corresponding threshold can be calculated directly. 

Consequently, while spikes are frequently generated during the neuron's initial convergence phase, the network activity subsides once a steady state is reached. New spikes are emitted only when the neuron's frequency shifts significantly, such as when a change in target range or velocity occurs within the radar signal. 
This event-based spiking mechanism is applied to both range and velocity neurons.

\begin{algorithm}[h]
\caption{Threshold-Based Spike Generation}
\label{alg:spike_generation}
\begin{algorithmic}[1]
\For{each neuron $n$}
    \State Update actual frequency $\omega_n$
    \State $\Delta \omega \gets \omega_n - \hat{\omega}_n$
    
    \While{$|\Delta \omega| > \theta$}
        \State Emit spike with polarity $\mathrm{sgn}(\Delta \omega)$
        \State $\hat{\omega}_n \gets \hat{\omega}_n + \theta \cdot \mathrm{sgn}(\Delta \omega)$
        \State $\Delta \omega \gets \omega_n - \hat{\omega}_n$
    \EndWhile
\EndFor
\end{algorithmic}
\end{algorithm}

\subsection{Summary of contributions}

In summary, we present a discrete-time adaptive resonate-and-fire (ARF) framework for FMCW radar signal processing based on adaptive-frequency neurons. In particular, we reformulate the continuous-time model proposed in \cite{buchli_frequency_2008} as a discrete-time dynamical system suitable for streaming signal processing and neuromorphic implementations.

Our approach adopts a complex-valued pure harmonic neuron/oscillator without nonlinear damping terms and extends the frequency adaptation rule to naturally support complex-valued radar signals. Furthermore, we use a discrete-time version of the feedback mechanism proposed in \cite{buchli_frequency_2008}, which allows multiple neurons to converge to different frequency components of the input signal. Importantly, compared to the original work, the proposed method adapts to both signal frequency and amplitude, allowing a single neuron to track each frequency component without requiring multiple neurons per target.

Finally, we propose a processing pipeline for both range and Doppler estimation using two layers of ARF neurons, together with an event-driven spike generation mechanism that emits spikes only when significant frequency changes occur, thereby reducing the communication bandwidth. 
Data and the code used in this study are available online\footnote{\url{https://github.com/TUE-EE-ES/adaptive_frequency_neurons_FMCW_radar}}.

\section{Results}
\label{sec:results}

\subsection{Synthetic data}

The following results were obtained by simulating a Frequency Modulated Continuous Wave (FMCW) radar system integrated with adaptive resonate-and-fire neurons. The simulation framework was developed in Python, using a C++ backend for high-performance numerical simulation of neurons. The radar parameters used throughout these experiments are detailed in Table \ref{tab:radar_params}. The FMCW simulation generates a standard radar frame of shape ($n_{\text{chirps}}$, $n_{\text{samples}}$). To allow processing by the neurons, this 2D frame is flattened into a continuous one-dimensional signal by stitching the chirps together. While the analysis in this section focuses primarily on the fast-time signal for range estimation, the underlying convergence principles are directly applicable to slow-time processing for velocity estimation.

We begin by evaluating the convergence behavior of a baseline case for a single-tone signal and a single neuron. Subsequently, we extend this analysis to more complex scenarios involving multi-frequency input signals.

\begin{table}[h]
\centering
\begin{tabular}{lll}
\hline
\textbf{Parameter} & \textbf{Description} & \textbf{Value} \\
\hline
$f_b$        & Carrier frequency        & $60\,\text{GHz}$ \\
$B$          & Bandwidth                & $2.0\,\text{GHz}$ \\
$n_{\text{chirps}}$  & Number of chirps        & $64$ \\
$n_{\text{samples}}$ & Samples per chirp       & $128$ \\
$t_{\text{chirp}}$   & Chirp duration          & $64\,\mu\text{s}$ \\
\hline
\end{tabular}
\caption{Synthetic Radar parameters used for the simulations.}
\label{tab:radar_params}
\end{table}

\subsubsection{Single target Analysis}

\begin{figure}[h]
    \centering
    \includegraphics[width=.6\linewidth]{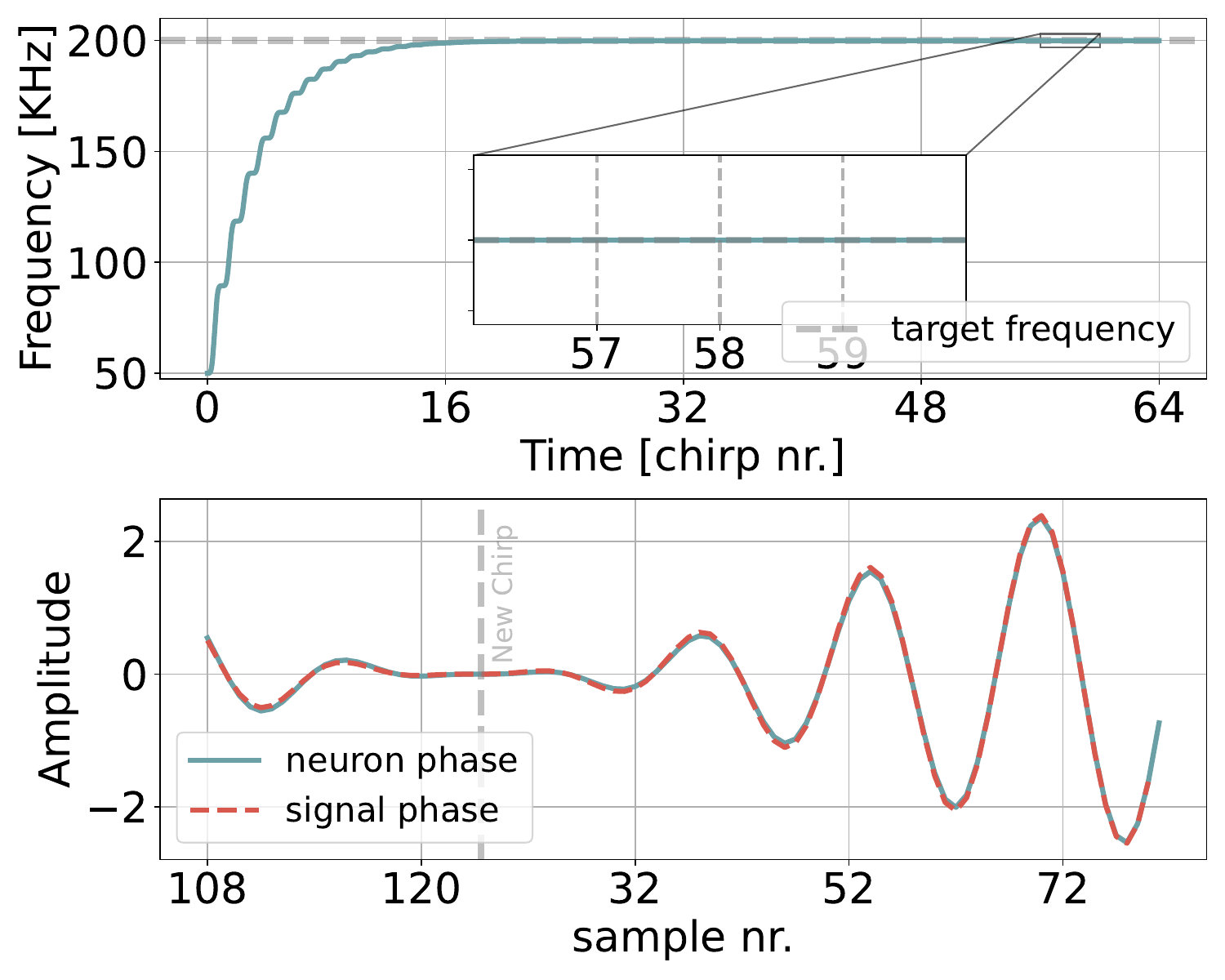}
    \caption{Frequency dynamics for a single neuron processing a single-tone input signal with a Hanning window. Top: Evolution of the neuron frequency over time starting from a random initial value. Bottom: Comparison of the windowed input signal (red), showing the amplitude change over time, and the converged neuron oscillation (blue) matching the input. The figure also shows the smoother transition between chirps resulting from applying a Hanning window.}
    \label{fig:single_oscillator_detail_hann}
\end{figure}

We start by analyzing the dynamics of a single neuron in the case of a single tone input signal. Figure \ref{fig:single_oscillator_detail_hann}(top) shows an example of the frequency of the neuron over time. The neuron starts with a random frequency and zero amplitude and after an initial adaptation period, the neuron's frequency converges to the target frequency. 

Figure \ref{fig:single_oscillator_detail_hann} (bottom) shows the input signal alongside the signal generated by the neuron. As mentioned earlier, a Hanning window has been applied to the input per chirp, in order to smooth the transitions between chirps.

\begin{figure}[t]
    \centering

    \begin{subfigure}[t]{0.48\textwidth}
        \centering
        \includegraphics[width=\linewidth]{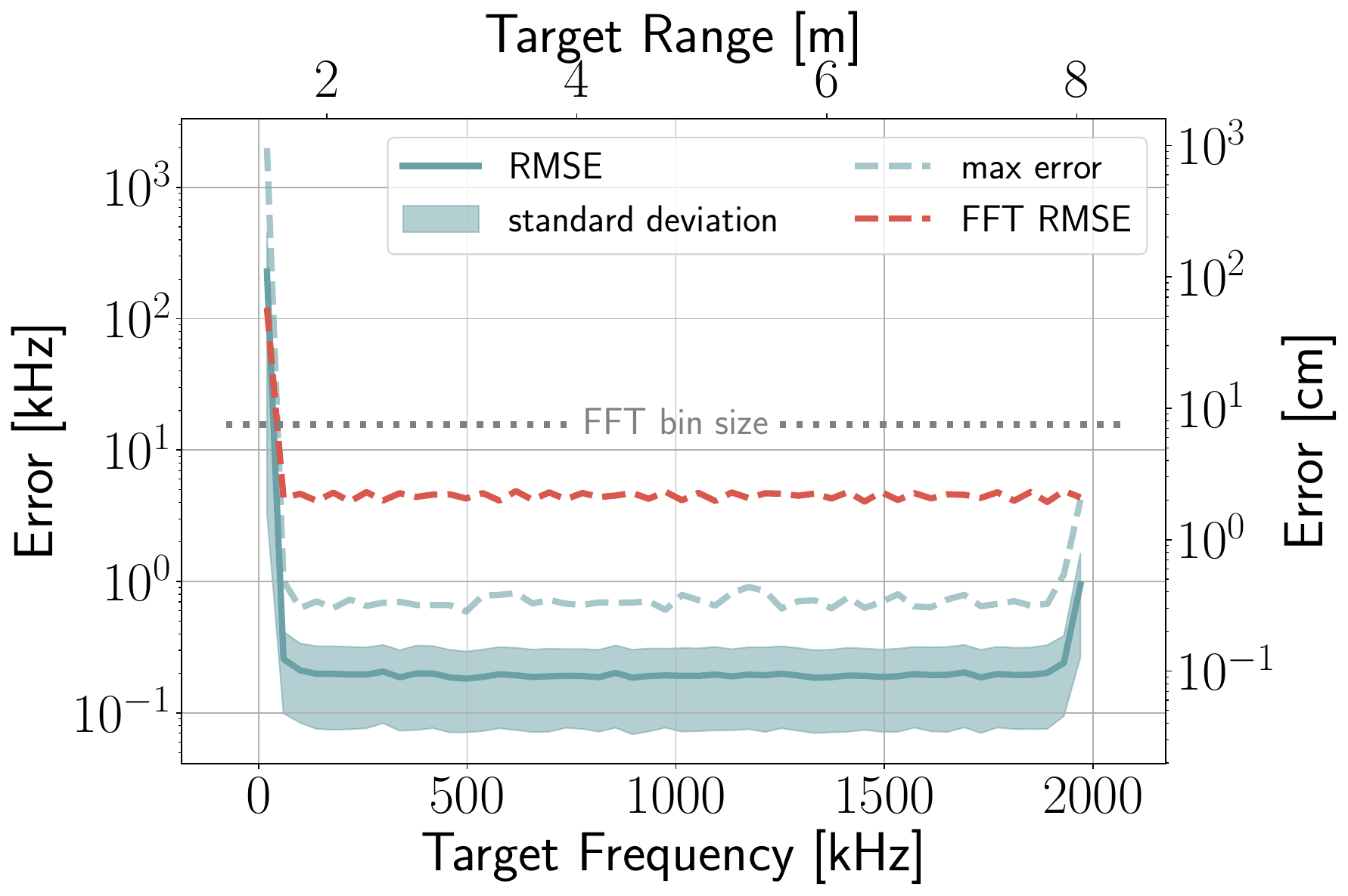}
        \caption{Error vs. target range.}
        \label{fig:subfig_range}
    \end{subfigure}\hfill
    \begin{subfigure}[t]{0.48\textwidth}
        \centering
        \includegraphics[width=\linewidth]{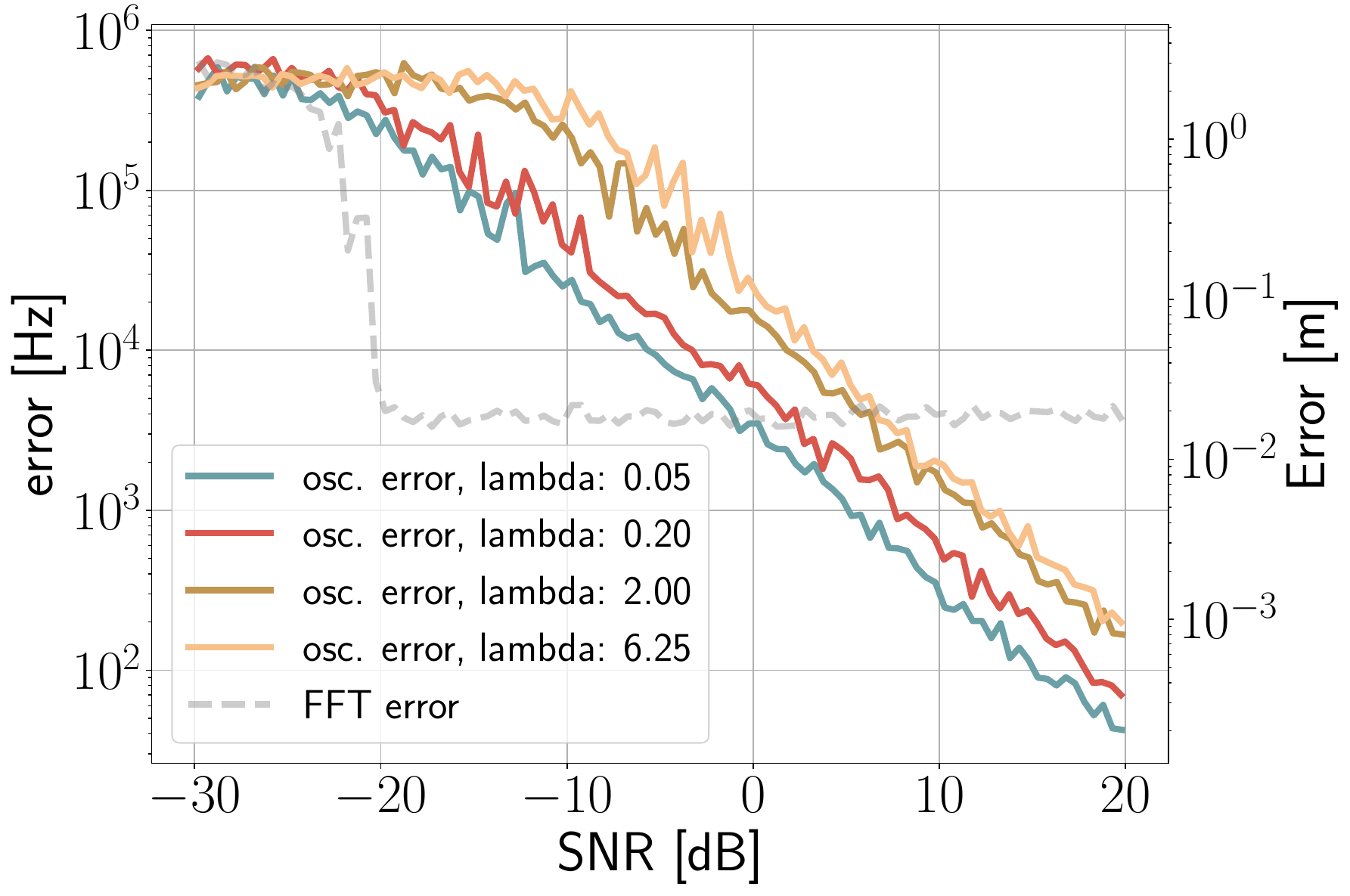}
        \caption{Error vs. input SNR.}
        \label{fig:subfig_snr}
    \end{subfigure}

    \vspace{0.8em}

    \begin{subfigure}[t]{0.48\textwidth}
        \centering
        \includegraphics[width=\linewidth]{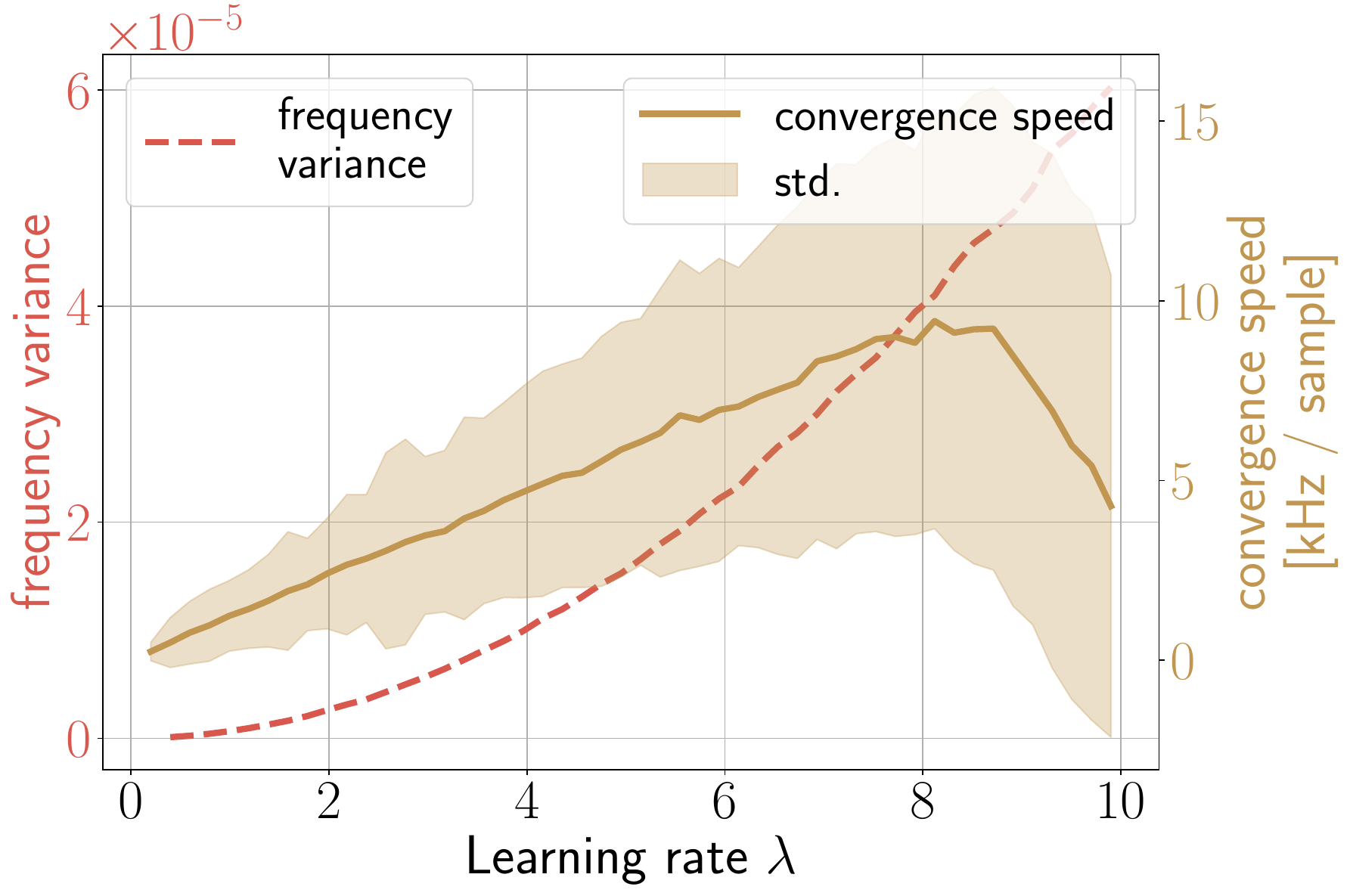}
        \caption{Impact of learning rate $\lambda$.}
        \label{fig:subfig_lambda}
    \end{subfigure}\hfill
    \begin{subfigure}[t]{0.48\textwidth}
        \centering
        \includegraphics[width=\linewidth]{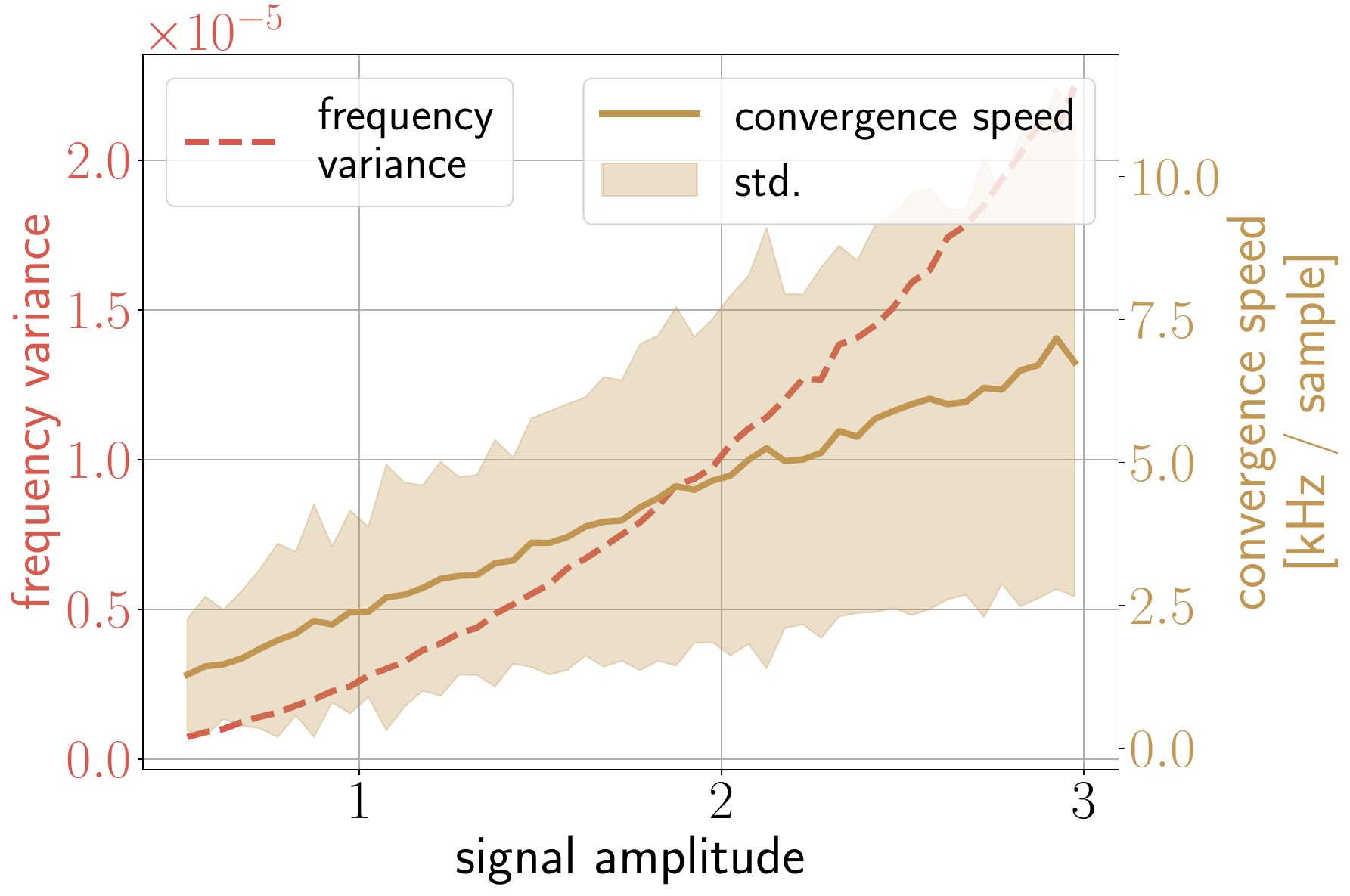}
        \caption{Impact of signal amplitude.}
        \label{fig:single_target_experiments_amplitude}
    \end{subfigure}

    \caption{Single-neuron performance metrics evaluated across 50,000 simulations with random target distances and initial neuron frequencies. Top row displays tracking accuracy binned by (a) target range and (b) input SNR level. Bottom row shows convergence speed and frequency stability binned by (c) learning rate $\lambda$ and (d) input signal amplitude.}
    \label{fig:single_target_experiments}
\end{figure}

We tested the convergence properties of the neuron under different conditions. We start by analyzing the convergence for different target frequencies. We run 50K experiments with random target frequency, random initial frequency of the neuron (both in the interval [0, 2000] kHz), random signal phase [-$\pi$, $\pi$], and random amplitude [.5, 3.0]. Then the statistics are drawn from all the simulations. For this specific test, we generated data with a SNR of \SI{20}{\decibel}. 

Figure \ref{fig:subfig_range} shows the error depending on the target frequency. Since complex IQ sampling is used, negative frequencies can arise; these are mapped to their corresponding non-negative frequencies for analysis. 
The results show that the neuron reliably converges to all target frequencies. Moreover, it achieves a lower estimation error than a traditional L-point FFT, where L is the number of samples. While comparable accuracy can be obtained with FFT-based methods through techniques such as zero-padding, our approach reaches similar performance in a single pass over the input signal and with minimal memory requirements.

We also evaluated the behavior of the model under different input noise levels. Figure \ref{fig:subfig_snr} compares the same model configuration, using complex input and a Hanning window, for different adaptation rates $\lambda$. As expected, larger values of $\lambda$ increase the sensitivity to noise but enable faster convergence, while smaller values reduce the sensitivity to noise at the cost of slower convergence. At low SNR levels, a traditional FFT performs significantly better. Note that for low values of $\lambda$, a single frame of data may not be sufficient for convergence. Hence, to evaluate the effect of noise, we used a sequence of 20 frames. This ensures that the error is measured once the neuron has converged to the correct frequency.

We analyze the effect of the adaptation rate $\lambda$ on both the stability of the neuron's frequency estimate and the convergence speed. The frequency stability is quantified as the variance of the estimated frequency at the end of the simulation. The convergence speed is defined as the frequency difference between the initial frequency of the neuron and the signal frequency divided by the time required for convergence. Figure \ref{fig:subfig_lambda} shows the results for different values of $\lambda$. As expected, lower values of $\lambda$ lead to slower convergence but improved stability, whereas higher values result in faster convergence. In low-noise scenarios, convergence can be achieved with only a few processed samples. When $\lambda$ becomes too large, the frequency estimate becomes unstable and the model fails to converge reliably. For these experiments, an SNR level of 20 dB was used.

Finally, figure \ref{fig:single_target_experiments_amplitude} shows the effect of signal amplitude on convergence speed and stability. Increasing the signal amplitude has a similar effect to increasing $\lambda$, because amplitude influences the phase measurement in Eq. \ref{eq:base_adaptive}. The neuron model could be modified to normalize both the neuron's state and the input; however, we found that this model offers the best results in terms of stability and reliability. We leave the study of alternative models for future work.

\subsubsection{Two Targets Analysis}

\begin{figure}[h]
    \centering
    \begin{subfigure}[t]{0.48\textwidth}
        \centering
        \includegraphics[width=\linewidth]{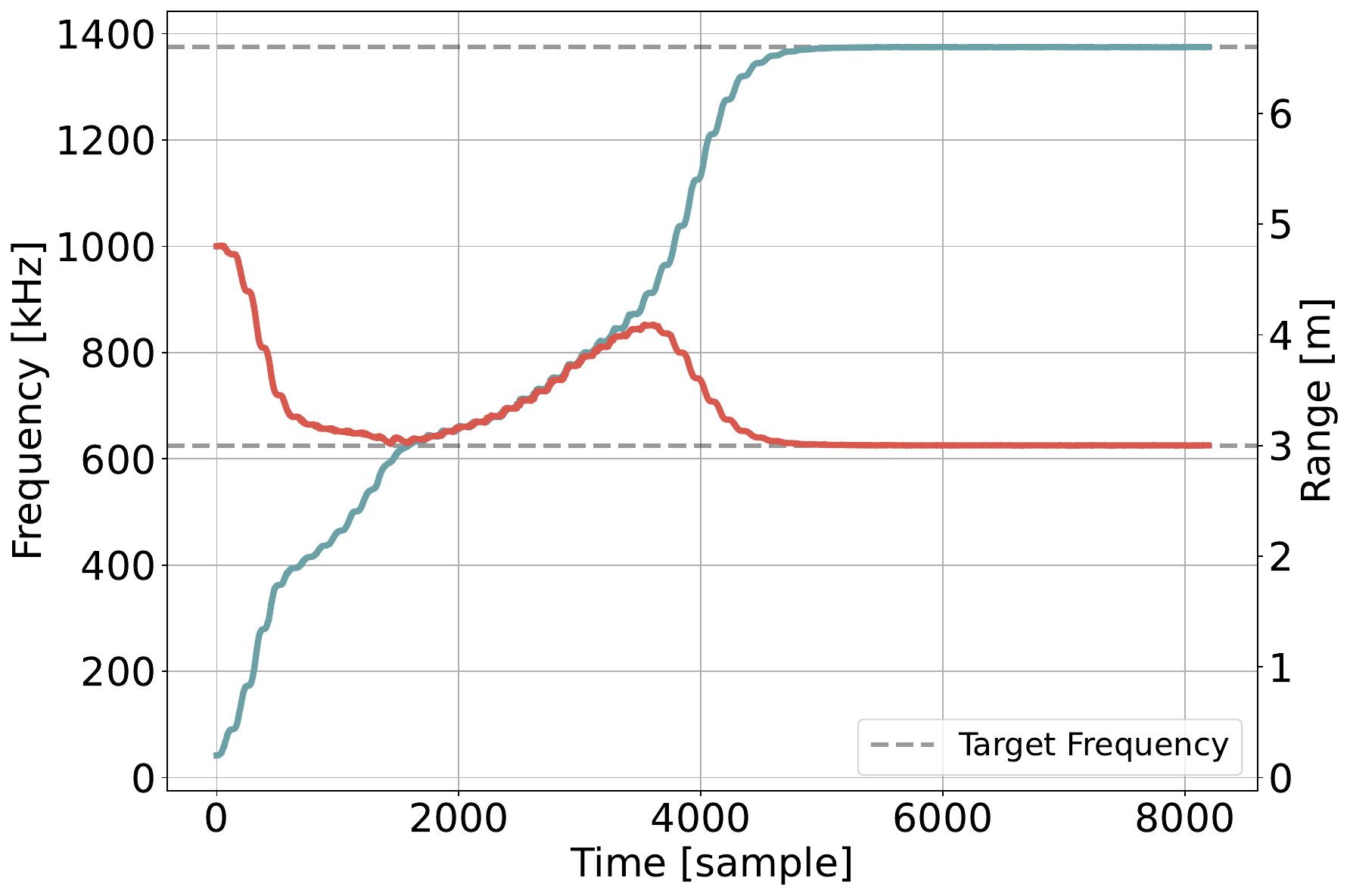}
        \caption{Two-target convergence.}
        \label{fig:two_oscillators_merge}
    \end{subfigure}\hfill
    \begin{subfigure}[t]{0.48\textwidth}
        \centering
        \includegraphics[width=\linewidth]{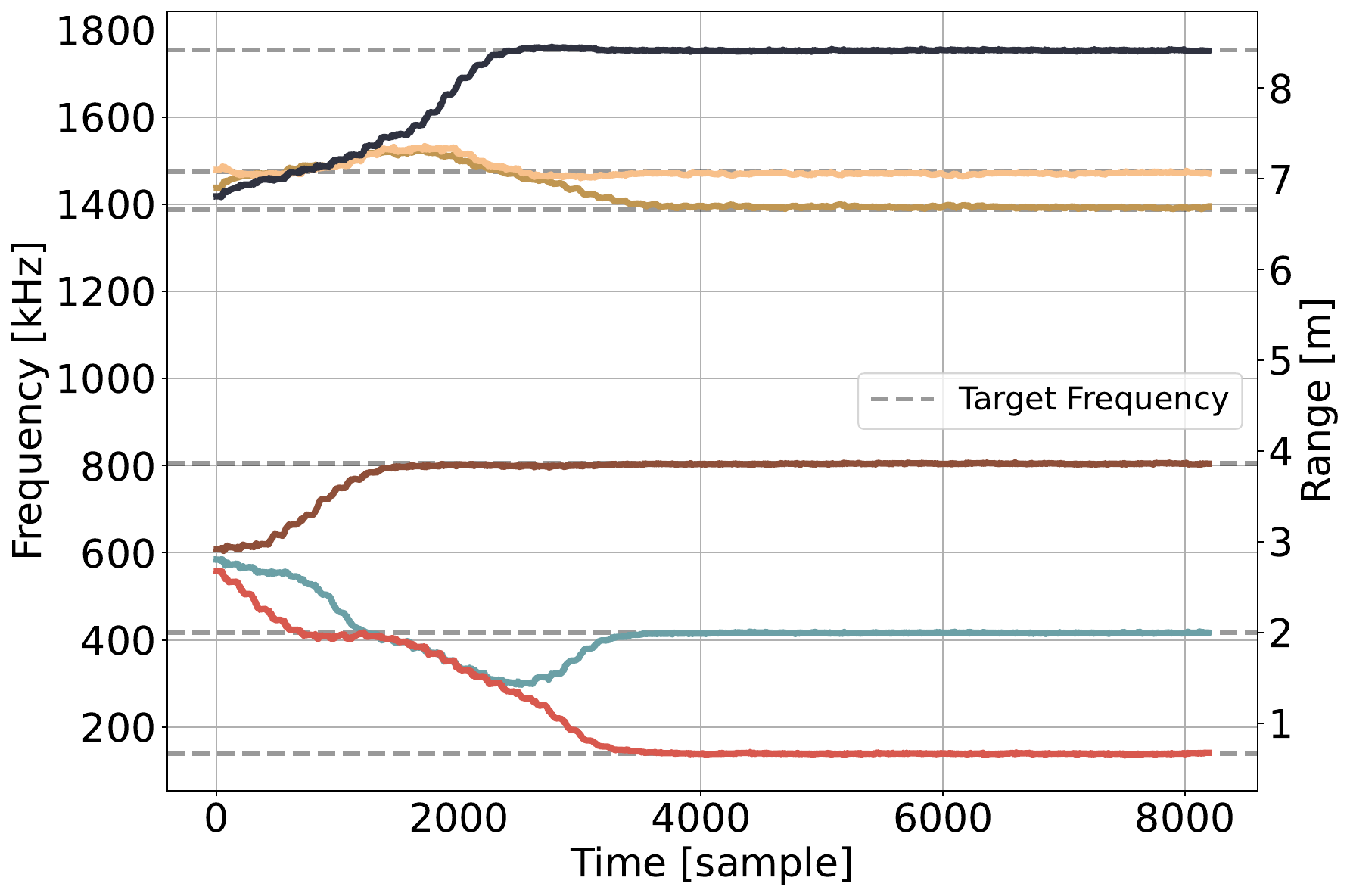}
        \caption{Six-target convergence.}
        \label{fig:six_oscillators_merge}
    \end{subfigure}
    \caption{Frequency dynamics for multiple neurons. (a) Simulation of two neurons settling onto a two-tone input signal. (b) Simulation of six neurons settling onto a six-tone input signal.}
    \label{fig:multi_target_merge_examples}
\end{figure}

\begin{figure}
    \centering
    \begin{subfigure}[t]{.45\textwidth}
        \centering
        \includegraphics[width=\textwidth]{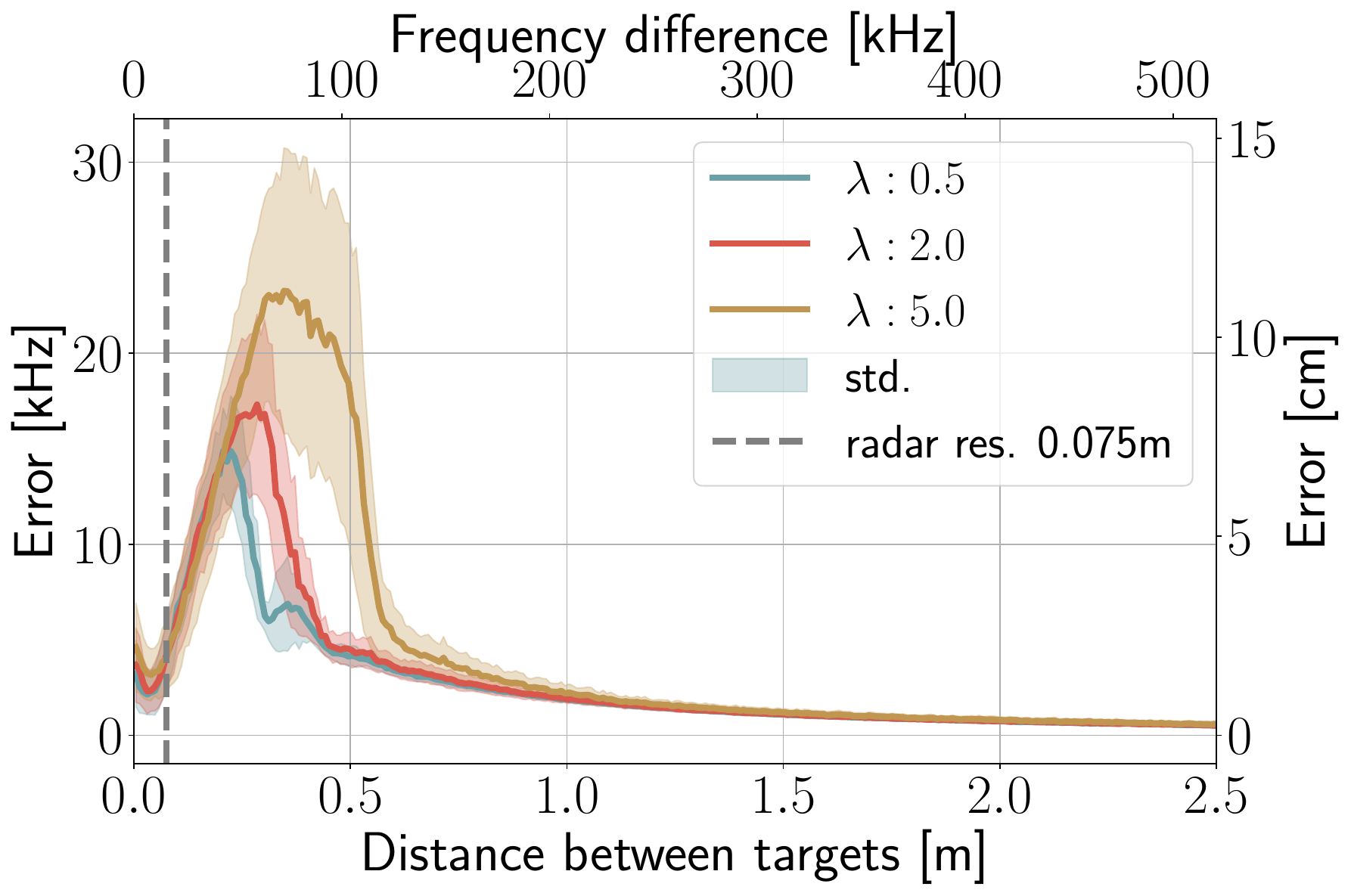}
        \caption{Convergence error vs. target distance.}
        \label{fig:oscillator_two_targets_failed_convergence}
    \end{subfigure}
    \hspace{1cm}
    \begin{subfigure}[t]{.45\textwidth}
        \centering
        \includegraphics[width=\textwidth]{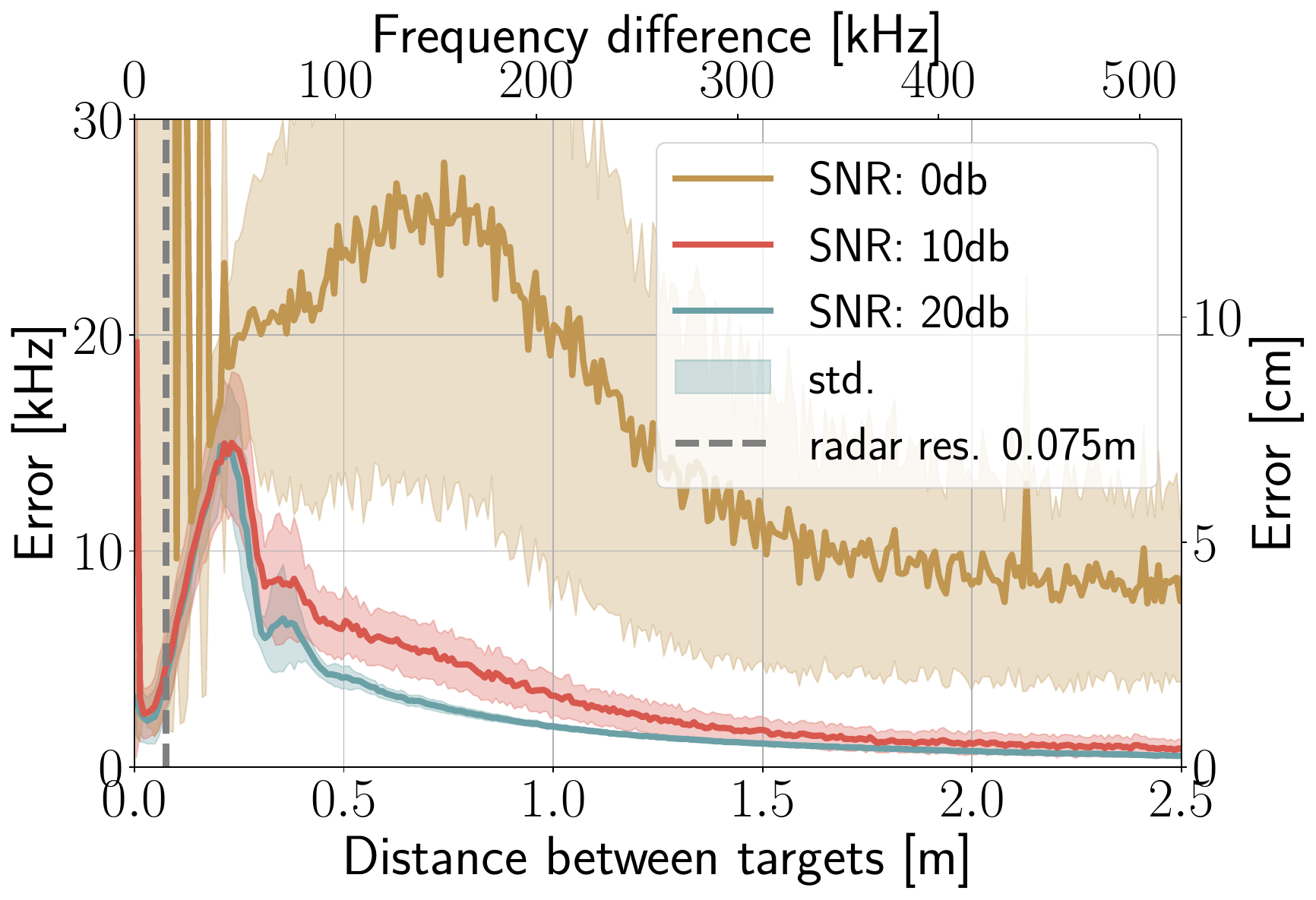}
        \caption{Convergence error vs. input SNR.}
        \label{fig:oscillator_two_targets_error_noise}
    \end{subfigure}

    \vspace{0.8em}

    \begin{subfigure}[t]{0.48\textwidth}
        \centering
        \includegraphics[width=\linewidth]{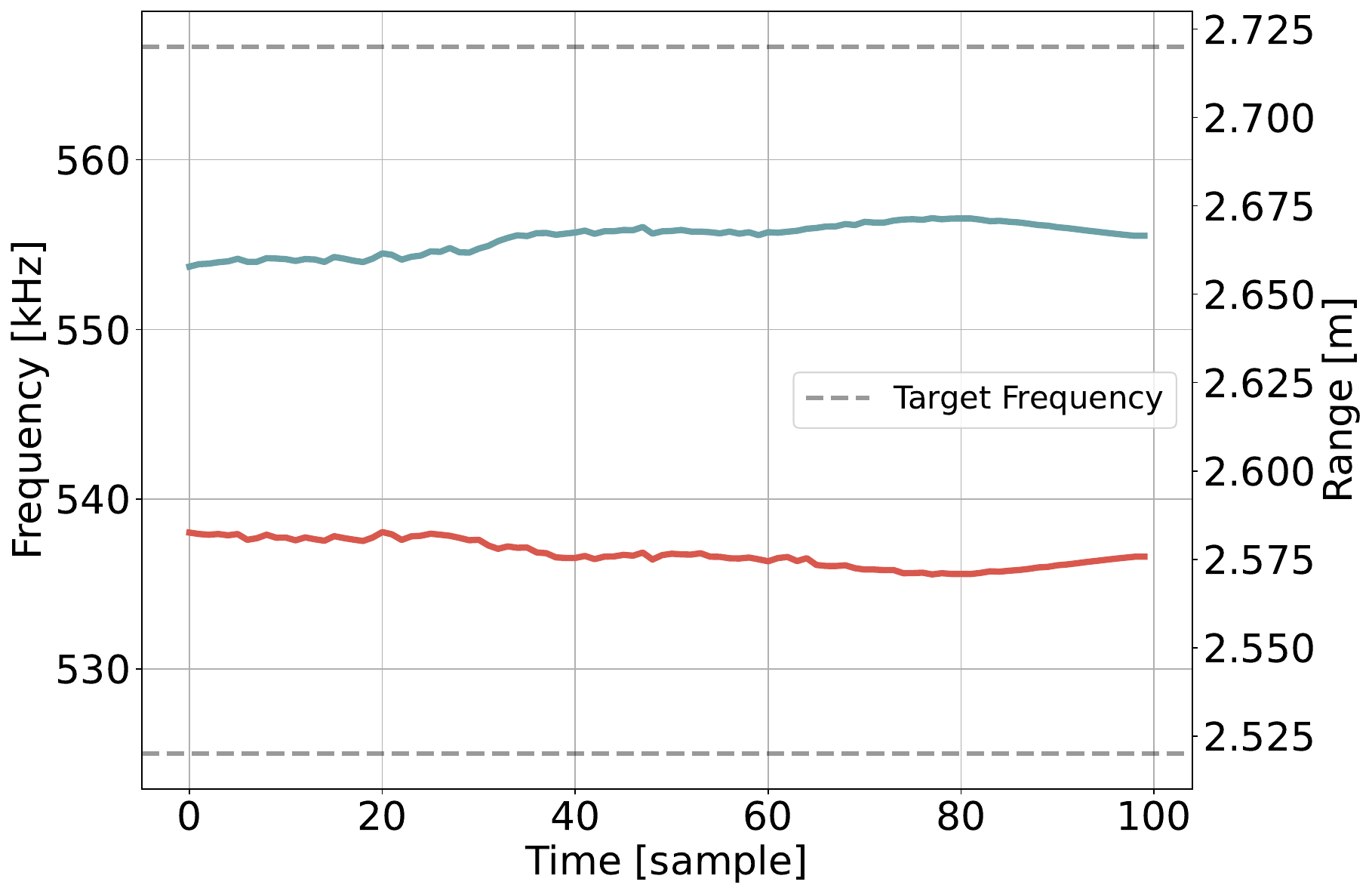}
        \caption{Inaccurate convergence for close targets.}
        \label{fig:oscillator_two_targets_error}
    \end{subfigure}\hfill
    \begin{subfigure}[t]{0.48\textwidth}
        \centering
        \includegraphics[width=\linewidth]{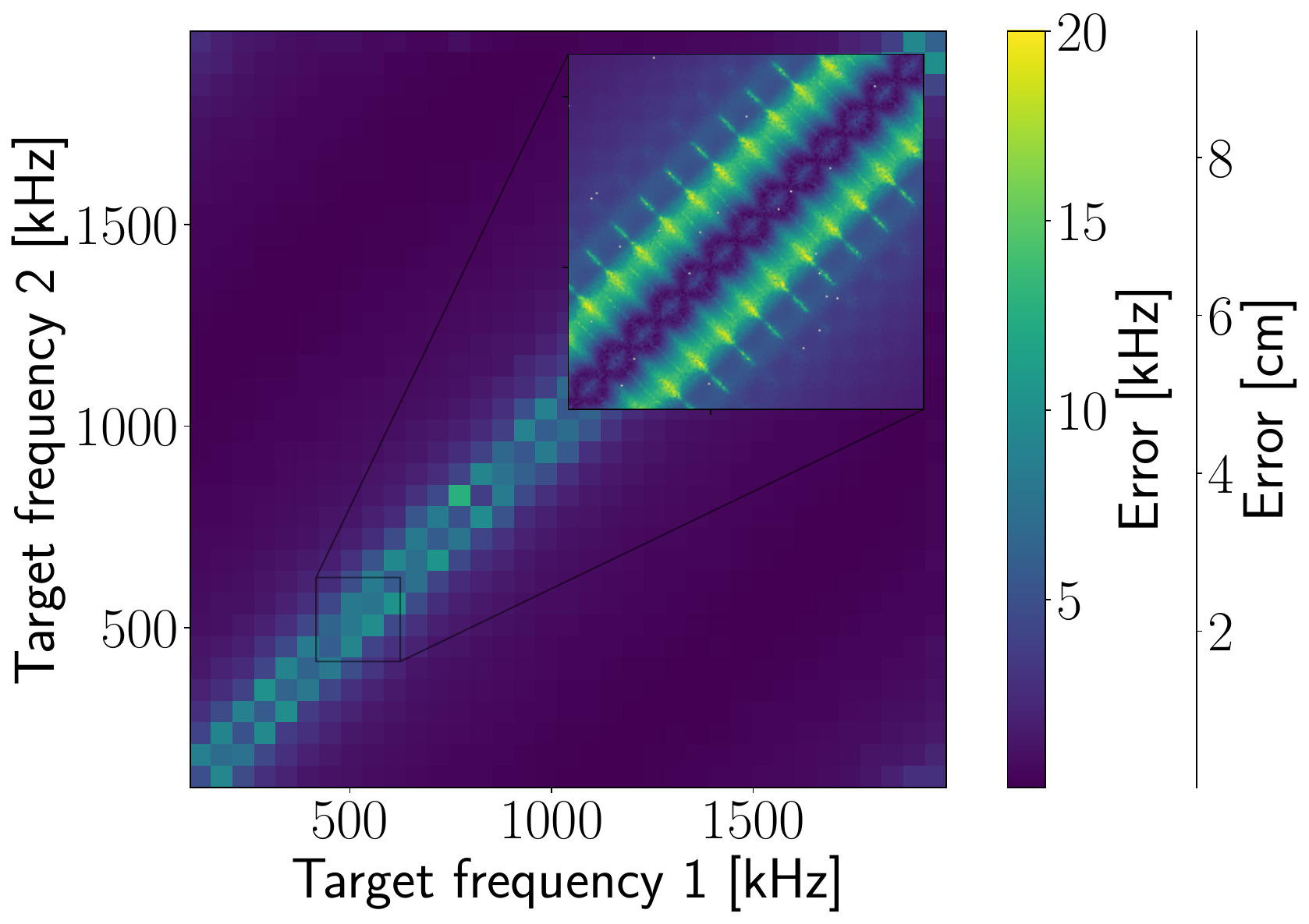}
        \caption{Convergence error map.}
        \label{fig:oscillator_two_targets_error_image}
    \end{subfigure}
    
    \caption{Two-target performance metrics using a two-tone input signal. (a) Total convergence error as a function of the distance between the two targets at a noise level of \SI{20}{\decibel}. (b) Total convergence error for different input SNR levels. (c) Example of neuron frequency evolution when two targets are close in frequency. The targets are too close and the neurons don´t converge precisely to the target frequencies.(d) Error map where each pixel represents a specific pair of target frequencies, with intensity indicating the final convergence error.}
    \label{fig:two_target_experiments}
\end{figure}

We now analyze the behavior of the system with two neurons and an input containing two frequency components. The feedback mechanism allows the neurons to converge to different frequencies. Figure \ref{fig:two_oscillators_merge} shows an example of the convergence behavior of two neurons. In this particular case, the neurons initially converge towards the same frequency and seem to be attracted to one another before eventually diverging towards the correct target frequencies. It is important to note that the effect of the feedback is always active. Even when the neurons have not yet converged to the target frequencies, they are still affecting the input signal through the feedback. This means that, during the convergence process, the neurons may introduce an additional frequency into the signal that is not present in the original input. This extra frequency can affect other neurons, which in turn also influence the original neuron. This may explain the unusual attraction observed between the neurons during the initial phase of convergence. Figure \ref{fig:six_oscillators_merge} the frequency convergence in the case of six neurons and six target frequencies. The same attraction phenomenon can be observed in this case as well. 

This recurrent interaction between neurons makes it very difficult to predict the exact behavior of the system, as in some cases, the neurons do not attract each other and instead converge immediately to the target frequencies. 
To study the behavior of the system, we again adopt a Monte Carlo simulation approach. We run 50K simulations with random initial conditions and then draw statistics from the recorded data.

Figure \ref{fig:oscillator_two_targets_failed_convergence} shows the total convergence error as a function of the distance between the two target frequencies. The plot shows that when the two targets are close together, the convergence error is higher. This is expected as two very similar frequencies are harder to distinguish. We tested the model with three different values of $\lambda$. From the figure, it is clear that slower frequency adaptation leads to a smaller overall error. Figure \ref{fig:oscillator_two_targets_error_noise} shows a similar plot, but with each line corresponding to a different level of input noise. As expected, higher noise levels reduce the accuracy of the results.

Figure \ref{fig:oscillator_two_targets_error} shows an example of convergence in which the two targets are only 20 cm apart. Note that for our radar parameters, the theoretical range resolution ($\Delta R = \frac{c}{2B}$) is \SI{7.5}{\cm}. Under these conditions, the two neurons cannot converge exactly to the target frequencies. Instead, they are pulled closer to each other. This is again likely related to the phenomenon discussed earlier, in which two neurons can attract each other during convergence. However, the behavior of neurons does not appear to be consistent across all frequencies. Figure \ref{fig:oscillator_two_targets_error_image} shows the convergence error as a function of the two target frequencies in the signal. It appears that some frequency combinations result in lower errors than others. 

\subsubsection{Spike Encoding}

\begin{figure}
    \centering
    \begin{subfigure}[t]{.49\textwidth}
        \centering
        \includegraphics[width=\linewidth]{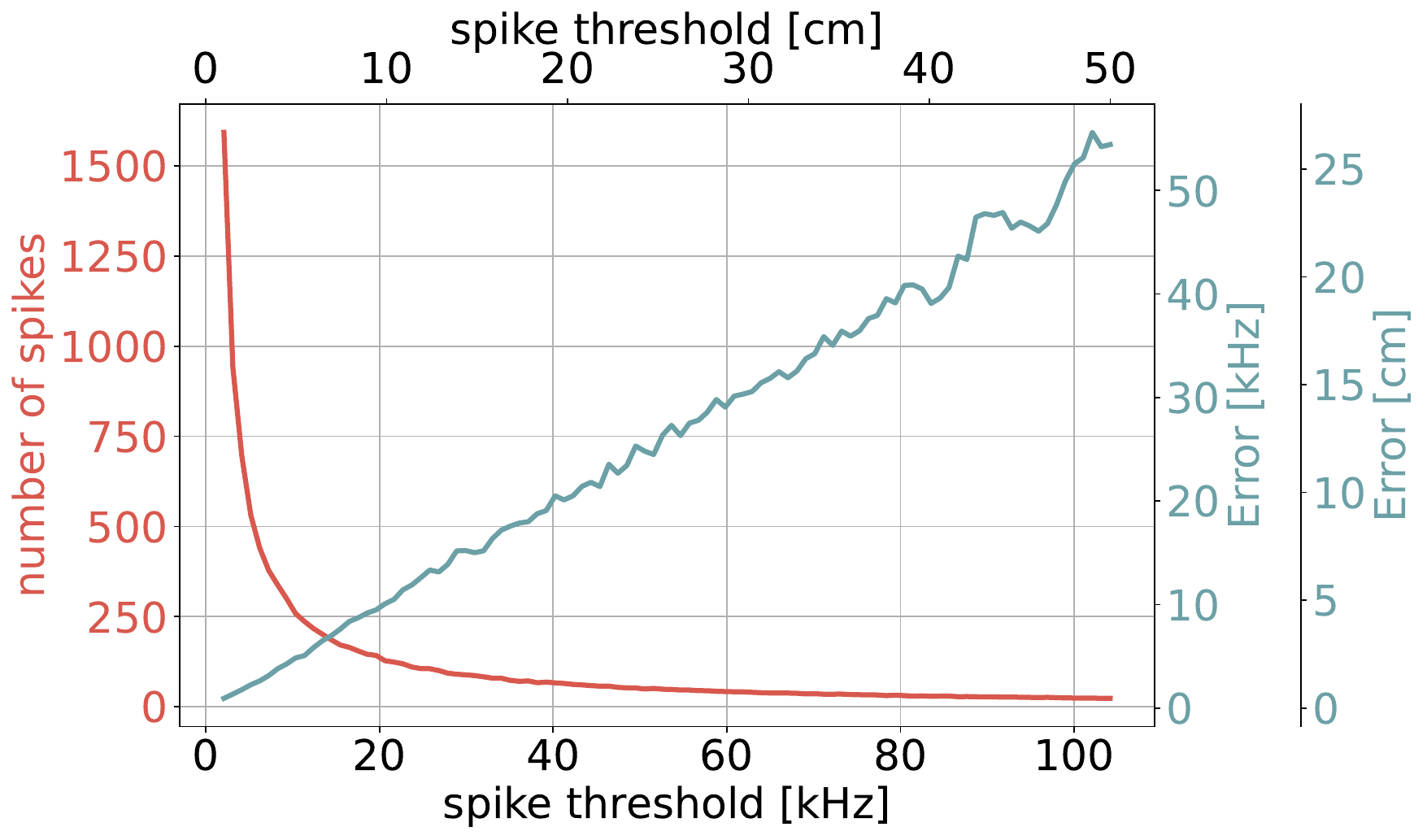}
        \caption{Spike count and reconstruction error.}
        \label{fig:spike_reconstruction_statistics}
    \end{subfigure}
    \begin{subfigure}[t]{.49\textwidth}
        \centering
        \includegraphics[width=\linewidth]{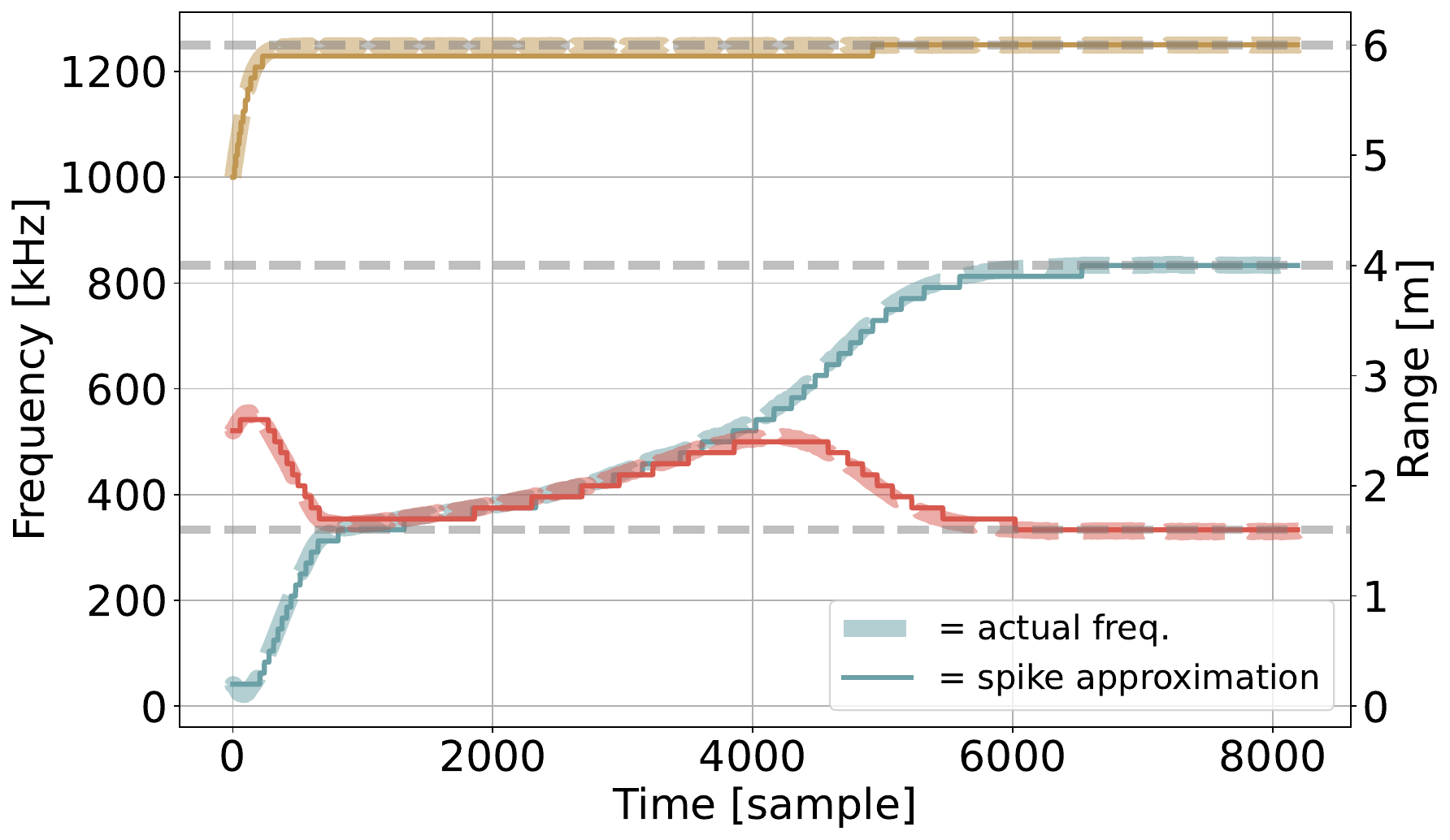}
        \caption{Frequency reconstruction from spikes.}
        \label{fig:spike_reconstruction_example}
    \end{subfigure}
    
    \caption{Spike output generation and reconstruction performance metrics. (a) Total number of spikes and the resulting signal reconstruction error evaluated across different values of the spiking threshold $\theta$, using 3 targets, 3 neurons, and an input SNR of approximately \SI{20}{\decibel}. (b) Comparison between the actual neuron frequency evolution over time and the frequency reconstructed solely from the emitted spikes, using a threshold of approximately \SI{20}{\kilo\hertz} (\SI{10}{\centi\meter}) which triggers a total of 75 spikes.}
    \label{fig:spike_statistics_and_example}
\end{figure}

The purpose of spike encoding is to reduce the communication bandwidth requirements between neurons. To evaluate its effectiveness, we measured the number of spikes generated for different threshold values. Figure \ref{fig:spike_reconstruction_statistics} illustrates how both the spike count and the reconstruction error vary with the spiking threshold. As expected, the results reveal a tradeoff between accuracy and communication efficiency: lower thresholds produce more spikes and higher accuracy, while higher thresholds reduce the number of spikes at the cost of increased reconstruction error. The figure also presents an example of comparing the reconstructed frequencies obtained from spikes with the ground-truth frequencies.

It is important to note that Figure \ref{fig:spike_reconstruction_statistics} reports the number of spikes generated during the convergence phase of the neurons. However, the main advantage of the spike encoding scheme emerges after convergence to the target frequency. Once converged, a neuron generates spikes only when a sufficiently large change in frequency is detected; otherwise, no spikes are produced. Consequently, the long-term spike rate depends strongly on the final application and on the dynamics of the target motion.

\subsection{Recorded Data}

\begin{figure}[t]
    \centering
    \begin{subfigure}[t]{.48\textwidth}
        \centering
        \includegraphics[width=\linewidth]{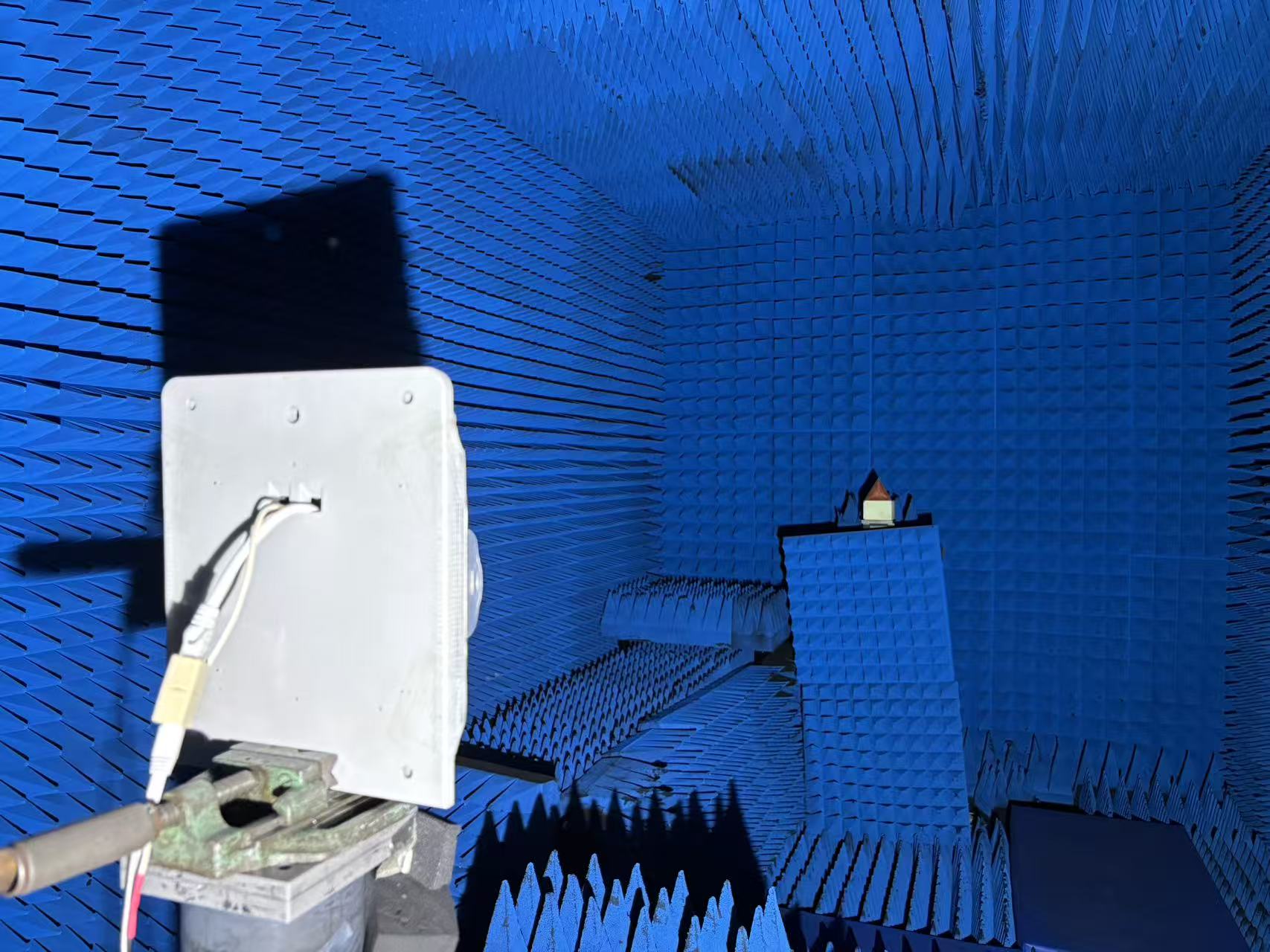}
        \caption{Anechoic chamber recording setup.}
        \label{fig:recording_setup}
    \end{subfigure}\hfill
    \begin{subfigure}[t]{.48\textwidth}
        \centering
        \includegraphics[width=\linewidth]{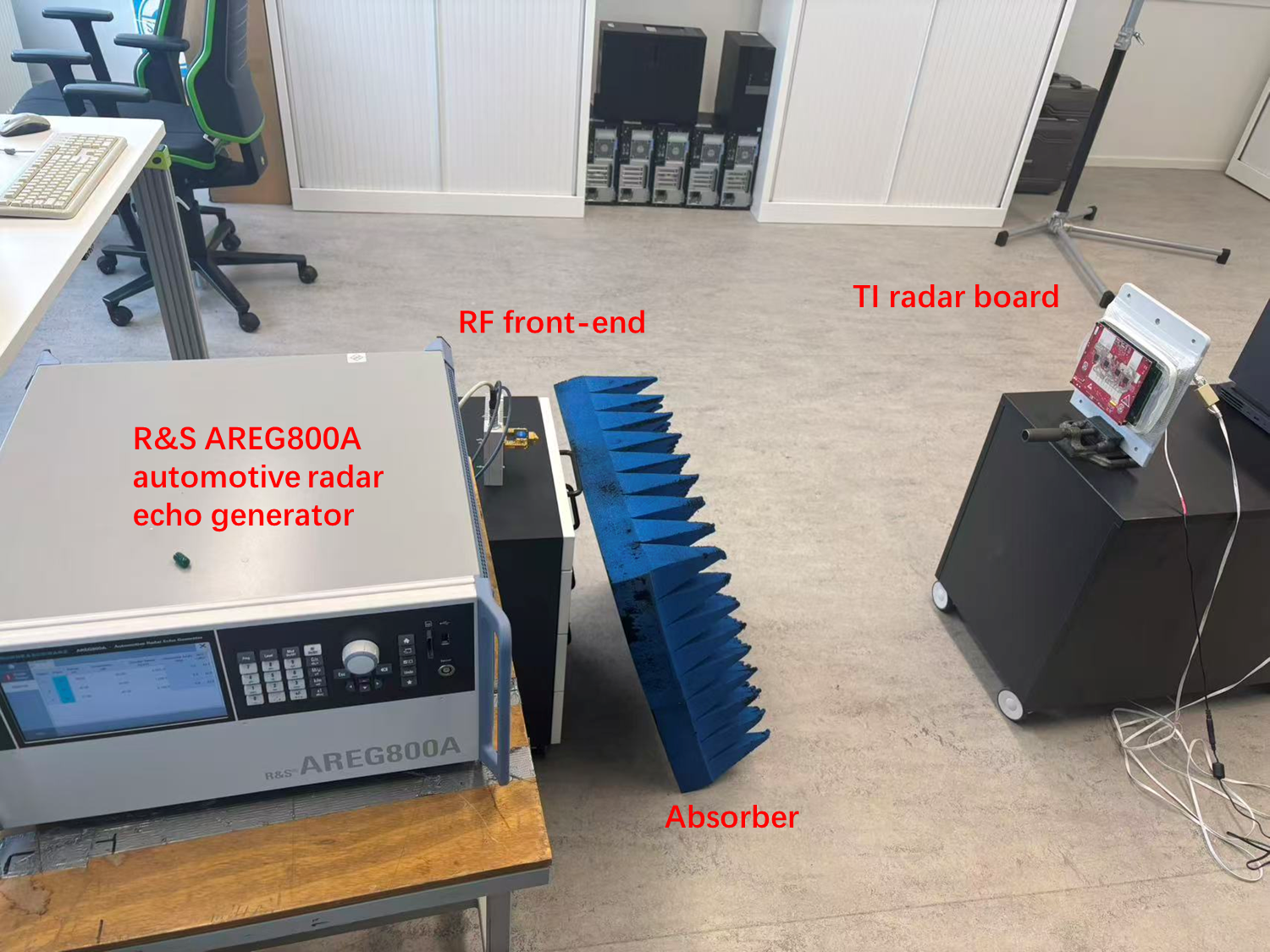}
        \caption{Automotive radar echo generator setup.}
        \label{fig:recording_setup3}
    \end{subfigure}
    \caption{Experimental hardware setups used for the data collection at TU Delft. (a) Single corner reflector measurement using the TI MMWCAS-RF-EVM cascade radar board within the DUCAT anechoic chamber. (b) Multi-target simulation configuration pairing the TI AWR2243 cascade radar with a Rohde \& Schwarz AREG800A automotive radar echo generator.}
    \label{fig:recording_setups}
\end{figure}

\begin{figure}[hbt]
    \centering
    \begin{minipage}[b]{0.48\textwidth}
        \centering
        \includegraphics[width=\linewidth]{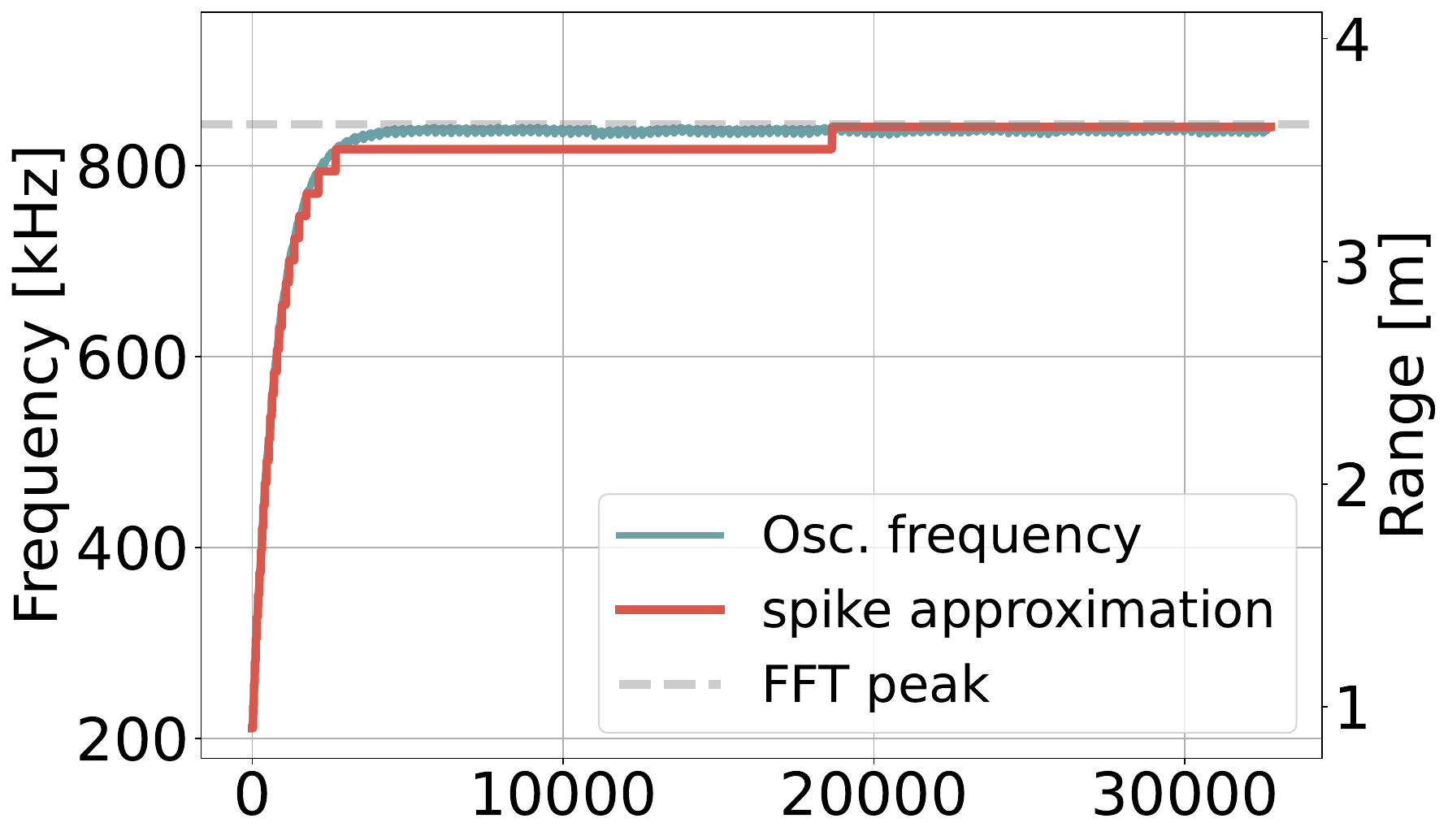}
        \caption{Frequency convergence dynamics evaluated on recorded hardware data.}
        \label{fig:single_target_recording}
    \end{minipage}\hfill
    \begin{minipage}[b]{0.48\textwidth}
        \centering
        \small
        \begin{tabular}{lc}
            \toprule
            \textbf{Metric} & \textbf{Avg. Value} \\
            \midrule
            FFT Peak     & \SI{3.649}{\meter} \\
            Exact Freq.  & \SI{3.640}{\meter} \\
            Recon. Freq. & \SI{3.602}{\meter} \\
            Spike Count  & 30 \\
            \bottomrule
        \end{tabular}
        \vspace{3em} 
        \captionof{table}{Averages calculated across three experimental runs at a fixed target distance using varying radar sampling frequencies.}
        \label{tab:single_target_recording_metrics}
    \end{minipage}
\end{figure}

\begin{figure}[t]
    \centering
    \begin{subfigure}[b]{\linewidth}
        \centering
        \includegraphics[width=\linewidth]{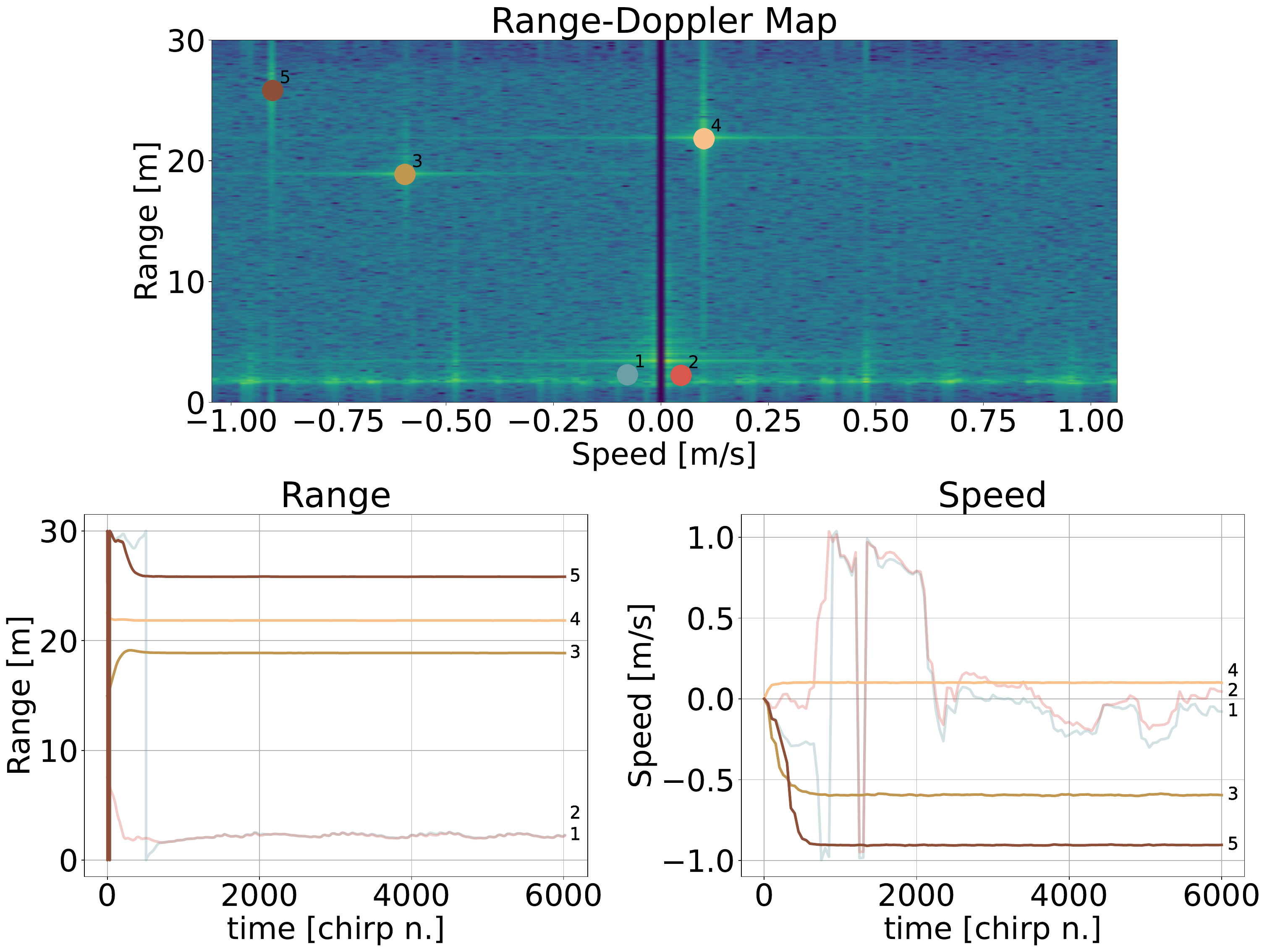}
        \caption{Range-Doppler map and frequency convergence of the neurons in terms of estimated range and velocity.}
        \label{fig:range_doppler_recorded_3_targets}
    \end{subfigure}
    
    \vspace{1em}
    
    \begin{subfigure}[b]{\linewidth}
        \centering
        \includegraphics[width=0.9\linewidth]{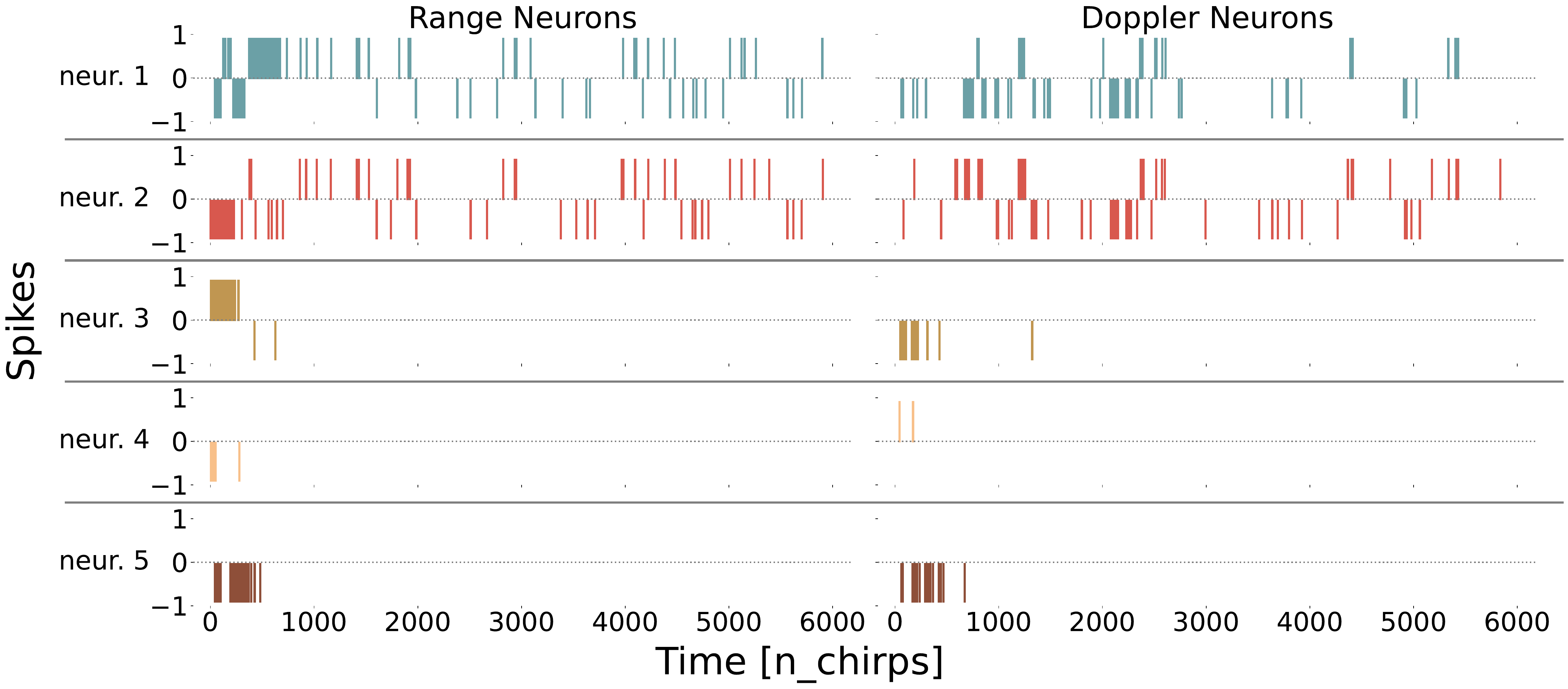}
        \caption{Spiking activity.}
        \label{fig:recorded_data_spike_raster}
    \end{subfigure}
    \caption{Multi-target performance on experimentally recorded data. (a) Top panel shows the Range-Doppler map overlaid with the final estimated steady-state neural states, indicated by the 5 circles. Bottom panels display the evolution of individual neural states over time for estimated range (left) and velocity (right). (b) Associated spike raster plot illustrating the spike times generated by the network for both range and velocity neurons.}
    \label{fig:multi_target_hardware_evaluation}
\end{figure} 

We first verified the range convergence using recorded data in a controlled environment. To evaluate the generalization capabilities of the adaptive neurons, we utilized a Texas Instruments (TI) MMWCAS-RF-EVM cascade radar board (AWR2243) \cite{Tidep} operating at 77 GHz. A single corner reflector was used as target within the DUCAT anechoic chamber at TU Delft, as depicted in Figure \ref{fig:recording_setup}. The experimental results align closely with those obtained from the synthetic data, demonstrating that the neuron can successfully converge to the target frequency. Figure \ref{fig:single_target_recording} illustrates the convergence dynamics of a single neuron on the recorded data, while the corresponding single-target performance metrics and total spike counts are detailed in Table~\ref{tab:single_target_recording_metrics}.

Furthermore, we evaluated the complete pipeline, encompassing both range and velocity estimation. For this evaluation, we employed the TI AWR2243 cascade radar board in conjunction with a Rohde \& Schwarz (R\&S) AREG800A automotive radar echo generator, as shown in Figure~\ref{fig:recording_setup3}. Three distinct targets were simulated using the generator, with detailed parameters provided in Table~\ref{tab:Simulation_three_tar}. The recorded data contains inherent environmental clutter, primarily manifesting as static targets with zero velocity, which challenges the convergence of the network. We demonstrate the convergence behavior on a dataset where this static clutter was removed; however, since some residual clutter remains, we found that 5 range neurons and 5 velocity neurons are required to ensure reliable convergence. While the network is capable of converging on raw data without prior clutter removal, doing so necessitates a significantly larger number of neurons (e.g. 15-20). Figure~\ref{fig:range_doppler_recorded_3_targets} illustrates the final converged values of the neurons and the associated frequency history. Table \ref{tab:radar_performance_comparison} shows the final estimated value for both range and velocity compared to the results obtained from an FFT. Finally, figure \ref{fig:recorded_data_spike_raster} shows the spikes generated by both the range and velocity neurons. The neurons that successfully converge to a target exhibit a stable frequency. Consequently, they generate spikes only during the initial convergence phase. Interestingly, the extra neurons that fail to converge to a specific target exhibit a much less stable frequency, resulting in a more continuous spike output. This could potentially be exploited to classify between converged and not converged neurons.

\begin{table}
\centering
\begin{tabular}{llll}
\hline
\textbf{Target indices} & Range (m) & Doppler speed (m/s)& Attenuation (dB) \\
\hline
1          & 18   & -0.6&-30\\
2        & 21   & 0.1&-40 \\
3          & 25   & -0.9&-35 \\
\hline
\end{tabular}
\caption{Simulation of three targets in the automotive radar echo generator.}
\label{tab:Simulation_three_tar}
\end{table}

\begin{table}
\centering
\begin{tabular}{lccccc}
\toprule
 & \multicolumn{2}{c}{\textbf{Range [m]}} & & \multicolumn{2}{c}{\textbf{Velocity [m/s]}} \\
\cmidrule{2-3} \cmidrule{5-6}
\textbf{Target} & \textbf{FFT + CFAR} & \textbf{ARF Estimate} & & \textbf{FFT + CFAR} & \textbf{ARF Estimate} \\
\midrule
Targ. 1 & 18.94 & 18.87 & & -0.59 & -0.59 \\
Targ. 2 & 21.88 & 21.83 & &  0.09 &  0.10 \\
Targ. 3 & 25.88 & 25.83 & & -0.91 & -0.90 \\
\midrule
\textbf{RMSE} & \textbf{0.9024} & \textbf{0.8483} & & \textbf{0.0073} & \textbf{0.0032} \\
\bottomrule
\end{tabular}
\caption{Comparison of range and velocity estimation performance for the conventional FFT-CFAR pipeline and the proposed ARF method. A systematic range offset of $\approx 0.8\,\text{m}$ is present in both methods due to the physical distance between the radar board and the simulator. For evaluation, values are rounded to two decimal places, and the RMSE is reported to four decimal places. To measure the estimation errors, correctly converged targets were manually selected from both the baseline CFAR outputs (which contain false-positive detections) and the ARF network (which utilizes two additional neuron pairs to ensure reliable convergence).}
\label{tab:radar_performance_comparison}
\end{table}

\subsection{Computational Complexity and Memory requirements}
A key advantage of the proposed method is its low memory usage. We compare our method against conventional methods such as the Discrete Fourier Transform (DFT) and Fast Fourier Transform (FFT), as well as prior neuromorphic implementations. Tables \ref{tab:complexity_comparison} and \ref{tab:complexity_comparison_neuromorphic} summarize the computational and memory complexity of these approaches.

Although the FFT is computationally efficient, it requires buffering the input signal before processing, which can lead to memory-access bottlenecks for large-scale problems. In contrast, our method processes the signal sequentially, sample by sample, without requiring signal buffering. The method only stores the internal states of the neurons, resulting in memory requirements that scale with the number of target frequencies. The computational complexity follows the same scaling, since each neuron is updated once per input sample, yielding the complexity reported in Table \ref{tab:complexity_comparison}.

The output of our method and the FFT are not directly equivalent. Rather than estimating the full frequency spectrum, the proposed method is designed to identify the dominant frequency components directly, thus incorporating also the process of detection in terms of radar processing.

Similarly, existing neuromorphic methods focus on replicating the mathematical operations of the DFT or FFT within a spiking framework \cite{lopez-randulfe_spiking_2021, lopez-randulfe_time-coded_2022}. As shown in Table \ref{tab:complexity_comparison_neuromorphic}, approaches like the Rate DFT and S-DFT have the $\mathcal{O}(L)$ memory complexity as a DFT, where $L$ is the signal length. This is because these methods typically require a neuron for each frequency bin $L$. While the S-FFT \cite{lopez-randulfe_time-coded_2022} reduces computational complexity to $\mathcal{O}(L \log L)$, it actually increases memory requirements and number of neurons.

In contrast, the memory footprint of our Adaptive RF method is decoupled from the spectral resolution $L$. Using the dynamical properties of ARF neurons, the system identifies and tracks only the dominant frequency components $N$. This eliminates the need for large input buffers as the signal is processed sample-by-sample. Consequently, the memory requirement scales only as $\mathcal{O}(N)$. In radar applications where the environment is sparse, this could allow for a significant reduction in the hardware resources required for real-time frequency analysis.

\begin{table}[h]
\centering
\begin{tabular}{lll}
\hline
\textbf{Method} & \textbf{Computational Complexity} & \textbf{Memory Footprint} \\
\hline
FFT          & $\mathcal{O}(L \log L)$   & $\mathcal{O}(L)$ \\
DFT          & $\mathcal{O}(L^2)$       & $\mathcal{O}(L)$ \\
\textbf{Adaptive RF} & $\mathcal{O}(N \cdot L)$ & $\mathcal{O}(N)$ \\
\hline
\end{tabular}
\caption{Computational and memory complexity for conventional methods. $L$ is the number of samples and $N$ is the number of ARF neurons.}
\label{tab:complexity_comparison}
\end{table}

\begin{table*}[t]
\centering
\setlength{\tabcolsep}{4pt}
\renewcommand{\arraystretch}{1.15}
\begin{tabular}{p{2.5cm} p{2.3cm} p{2.0cm} p{2.8cm} p{2.4cm} p{2.0cm}}
\hline
\textbf{Method} 
& \textbf{Input Type} 
& \textbf{Neuron Number} 
& \textbf{Number Spikes} 
& \textbf{Comp. Complexity} 
& \textbf{Mem. Footprint} \\
\hline
Rate DFT \cite{lopez-randulfe_spiking_2021} 
& s. mean-rate
& $2L$ 
& High
& $\mathcal{O}(L^2)$
& $\mathcal{O}(L)$ \\

S-DFT \cite{lopez-randulfe_time-coded_2022}         
& s. TTFS   
& $2L$ 
& $L \cdot 2L + 2L$
& $\mathcal{O}(L^2)$
& $\mathcal{O}(L)$ \\

S-FFT \cite{lopez-randulfe_time-coded_2022} 
& s. TTFS   
& $2L \log_{4}(L)$ 
& $8 \cdot 2L \log_4(L) + 2L$
& $\mathcal{O}(L \log L)$
& $\mathcal{O}(L \log L)$ \\

RF \cite{hille_resonate-and-fire_2022}
& Raw signal       
& $L$ 
& n.a.
& $\mathcal{O}(L^2)$
& $\mathcal{O}(L)$ \\

SpiNR \cite{reeb_energy-efficient_2026}
& Raw signal       
& $L$ 
& n.a.
& $\mathcal{O}(L^2)$
& $\mathcal{O}(L)$ \\

\textbf{Adaptive RF}
& Raw signal       
& $N$  
& n.a.
& $\mathcal{O}(N \cdot L)$
& $\mathcal{O}(N)$ \\
\hline
\end{tabular}
\caption{Computational and memory complexity for neuromorphic methods compared to the proposed Adaptive RF. $L$ is the number of samples and $N$ is the number of ARF neurons. }
\label{tab:complexity_comparison_neuromorphic}
\end{table*}

\section{Discussion and Outlook}
\label{sec:discussion}

The experiments validate the three-part hypothesis that motivates this work. Single-target simulations show that an ARF neuron converges onto the beat frequency of an isolated reflector with accuracy comparable to an FFT while storing only its own state; multi-target simulations show that the mean-field feedback partitions neurons onto distinct peaks; and recorded measurements confirm that the same primitive tracks physical targets under static and moving conditions. In all cases, the neuron count equals the number of tracked targets rather than the spectrum resolution, and the pipeline produces peak estimates directly, without a separate CFAR detection stage. 

That said, the present study can be significantly extended. This work presents a pure algorithmic contribution: we report no on-chip timing or energy numbers, so the quantitative comparison with digital neuromorphic pipelines such as SpiNR--AGS on Loihi~2~\cite{reeb_energy-efficient_2026} is at the complexity-class level rather than at the joule level. Convergence depends on the adaptation rate $\lambda$ and degrades under low SNR or when multiple targets fall within a sub-bin distance of one another; the mean-field feedback scheme also assumes that the number of neurons is at least equal to the number of dominant components. Finally, the recorded-data experiments are acquired in a controlled indoor setting and do not yet stress the method against more complex automotive clutter, or even interference.

Future work will aim to expand the current framework in several directions. First, hardware deployment: the online processing and dynamical nature of ARF neurons seem a good fit for analog resound-and-fire implementations~\cite{nakada_analog_2006,lehmann_direct_2023,liu_linear_2025}, where continuous-time oscillatory behavior can be realized efficiently. Digital hardware is also compelling, and we plan implement the current methods on digital off-the shelf hardware as FPGA or digital neuromorphic systems as Intel Loihi2 \cite{davies_advancing_2021}. This would enable a direct head-to-head energy comparison against SpiNR--AGS~\cite{reeb_energy-efficient_2026} on identical applications. Additionally, our objective is to further expand on the robustness and reliability of the framework, for example, currently we assume that the number of targets is known, but a better solution would be to dynamically allocate neurons as needed. Lastly, we plan to add a third ARF layer applied across antennas to recover angle-of-arrival by the same mechanism that handles range and Doppler, and the target-indexed output is already in a form well suited to downstream tracking and classification.

The proposed method constitutes a shift in the approach to radar signal processing, where the goal is to find a sparse frequency representation using a dynamical system. We believe that this method is well suited for radar applications with a limited number of targets. Furthermore, the underlying architecture can be generalized to other domains where extracting primary spectral components is more important than computing a full frequency decomposition.

\section{Conclusion}
\label{sec:conclusion}

In this paper, we introduced adaptive resonate-and-fire (ARF) neurons as a target-indexed, streaming alternative to FFT- and bin-based spiking radar pipelines. Unlike methods that allocate resources per frequency bin, a single ARF primitive, applied once along fast time and once along slow time, estimates range and Doppler directly from the dominant signal components, processing the input sample-by-sample without buffering and without a separate CFAR detection stage. Memory therefore scales with the number of tracked targets rather than with the spectral resolution, and the same primitive can replace any sparse-FFT block in which only a few dominant frequencies matter. 
Experiments on synthetic and experimentally recorded data confirm that the method reliably recovers single and multiple targets under static and moving conditions. The proposed discrete-time model offers significant reductions in memory requirements compared to Fourier-based approaches, and it is compatible with existing radar systems and digital neuromorphic hardware. Furthermore, because the formulation is adaptable to a continuous-time oscillatory dynamical system, analog neuromorphic implementations are a natural next step.

\section*{Acknowledgment}
The authors thank the Dutch Research Council (NWO) for funding this research with grant OCENW.M.22.331 (NERD).

\clearpage
\printbibliography

\end{document}